\documentclass[journal]{IEEEtran}
\usepackage{cite}
\usepackage{support-caption}
\usepackage{subcaption}
\usepackage[utf8]{inputenc}
\usepackage{caption}
\usepackage{multirow}
\usepackage{blindtext}
\usepackage{tabularx}
\usepackage[table,xcdraw]{xcolor}
\usepackage{array}
\usepackage{url}
\usepackage{color}
\usepackage{booktabs}
\usepackage{times}
\usepackage{soul}
\usepackage{url}
\usepackage[hidelinks]{hyperref}
\usepackage{amsmath}
\usepackage{graphicx}
\usepackage{xspace}
\usepackage{xcolor}
\usepackage{fancyhdr}
\usepackage{graphics}
\usepackage{makecell}
\usepackage{csquotes}
\usepackage{xr}
\usepackage{tikz}
\usepackage{pgfplots}
\pgfplotsset{compat=1.16}
\usepackage{wrapfig}
\usepackage{enumitem}
\usepackage{ulem} 
\usepackage{balance}
\usepackage[T1]{fontenc}
\definecolor{green}{rgb}{0.0, 0.5, 0.0}
\definecolor{amethyst}{rgb}{0.6, 0.4, 0.8}

\begin{document}
\title{ViT-ReT: Vision and Recurrent Transformer Neural Networks for Human Activity Recognition \\in Videos}

\author{James~Wensel,
        Hayat~Ullah,~\IEEEmembership{Student Member,~IEEE,} and
        Arslan~Munir,~\IEEEmembership{Senior Member,~IEEE}
\thanks{
This research was supported in part by the Air Force Office of Scientific Research (AFOSR) Contract Number FA9550-22-1-0040. (Corresponding author: Arslan Munir)
\newline
\indent J. Wensel, H. Ullah and A. Munir are with the Intelligent Systems, Computer Architecture, Analytics, and Security Laboratory (ISCAAS Lab), Department of Computer Science, Kansas State University, Manhattan, Kansas, 66506, USA (e-mail: \href{mailto:}{jdwensel@ksu.edu}, {hayatu@ksu.edu}, \href{mailto:}{amunir@ksu.edu}).\\
}
}
\maketitle

\begin{abstract}
Human activity recognition is an emerging and important area in computer vision which seeks to determine the activity an individual or group of individuals are performing. The applications of this field ranges from generating highlight videos in sports, to intelligent surveillance and gesture recognition. Most activity recognition systems rely on a combination of convolutional neural networks (CNNs) to perform feature extraction from the data and recurrent neural networks (RNNs) to determine the time dependent nature of the data. This paper proposes and designs two transformer neural networks for human activity recognition: a recurrent transformer (ReT), a specialized neural network used to make predictions on sequences of data, as well as a vision transformer (ViT), a transformer optimized for extracting salient features from images, to improve speed and scalability of activity recognition. We have provided an extensive comparison of the proposed transformer neural networks with the contemporary CNN and RNN-based human activity recognition models in terms of speed and accuracy.
\end{abstract}
\IEEEpeerreviewmaketitle
\section{Introduction}
\IEEEPARstart{O}{ne} of the most important and fastest growing fields in machine learning is computer vision. Computer vision is the process of using a computer to interpret the real world through some form of sensor. Modern computer vision involves using artificial neural networks to determine patterns or features in the data from the sensor and then using these discovered features to inform other processes. This process can be used to learn from images or videos by treating each frame in the video as a separate image and performing processing on each frame and the video as a whole. In recent years, convolutional neural networks (CNNs) have gained significant traction in the computer vision community for image classification \cite{CiresanClassification}. Popular CNN models used for image classification, such as AlexNet, VGG-16, and ResNet contain long chains of convolutional layers of different densities and filter sizes along with pooling and activation layers. As early as 2012, this task led to very deep implementations of image classifiers \cite{KrizhevskyImageNet}, and even more recently, deep CNNs with residual connections that provided even better results \cite{HeDeepResL}. This final approach created ResNet \cite{HeDeepResL}, a residual network which uses a residual connection to recombine the extracted features with previous inputs at different time steps in the model. This approach helped create more robust models that could be trained more easily than deep CNNs without residual connections \cite{HeDeepResL}. These models, once trained, can even be transferred to domains they were not initially trained in, via a technique called transfer learning \cite{AmosTransfer}. This process has been done with relative success depending on both the dataset used for initial training and the domain of the testing data \cite{TimmeCNN}. CNN models, including the ResNet classifier, have seen recent success and have been the dominant models used for computer vision tasks. However, the high model complexity of deep CNNs \cite{StriglScalability} is not ideal for all systems and shows a potential need for new model architectures. For example, for the internet of things (IoT), a network of physical devices which generates data from multiple different sources \cite{MunirIFCIoT}, there is a need to process the generated data efficiently. In many of these scenarios, the devices have limited resources to apply to the image classification or object detection tasks. Furthermore, newer trends in computing, such as edge computing \cite{MunirDataFusion}, where the data is processed closer to the data source as opposed to cloud computing paradigm, necessitate more efficient and lightweight implementations. Consequently, there is a pressing need for developing lightweight human activity recognition models that can be executed efficiently on resource constrained IoT and edge devices.\\
\indent Human activity recognition is a more complex task than image classification or object detection. Often the activities that are being classified involve some form of time dependence and cannot be determined from a single frame. For example, an alley-oop dunk, one of the most exciting highlights in all of sports, consists of the combination of a drive from a player, a pass to said player, a catch, and a dunk. There are few points, if any, at which a single frame of this sequence would give enough information to determine the action being performed, making it hard to classify this highlight. This is similarly true in soccer \cite{MunirSoccer}, as well as non-sports domains such as intelligent surveillance \cite{QinActionSimilarity} where most actions consist of a combination of smaller actions that must all be similarly classified. These classifications are then combined to determine the overall action from the data. This is done by combining the features extracted by a CNN with the time-dependent feature analysis of a recurrent neural network (RNN).\\
\indent The RNN maintains hidden states that allow for greater impact of time dependencies on their input data \cite{GravesSpeech}. Many variants of RNNs exist, with the two most used in activity recognition being long short-term memory (LSTM) units, and gated recurrent units (GRUs). LSTMs have an internal hidden layer that contains multiple memory cells to “remember” previous inputs. These cells inform three different output gates which serve as the output and updaters for future LSTMs \cite{OjoStocks}. GRUs are much the same but maintain fewer internal states and have two output gates instead of three \cite{ChoEncDec}. However, GRUs maintain similar performances to LSTMs \cite{HeckSimpGate}. While RNNs have seen some success, they also suffer from the same complexity issues as CNNs. RNNs are also not obviously parallelizable, as an RNN layer contains multiple RNN units that compute the output of the layer sequentially. As a result, RNNs often have long training and prediction times \cite{ViswaniAttention}. This can lead to issues when operating in real-time systems or on edge devices. To fix this, “attention mechanism” were constructed, which allow the model to “pay attention” to certain parts of the input and ignore others. Attention was first used in conjunction with RNNs in sequence-to-sequence models but was later used on its own in development of the transformer \cite{ViswaniAttention}. The transformer was shown to not only be more efficient and lightweight, but to have similar performance to traditional methods when trained on a sufficiently large dataset. Our main contributions in this paper are as follows:

\begin{enumerate}
    \item Elaborating and exploring the functionality and effectiveness of the transformer neural networks (TNNs).
    \item Presenting the domain adoption framework showing the applicability of TNNs for human activity recognition task.
    \item Proposing a recurrent TNN (ReT) to replace the computationally complex RNN in the typical activity recognition network chain.
    \item Proposing a specialized vision TNN (ViT) to replace the computationally complex CNN feature extractor in the typical activity recognition network chain. 
    \item Evaluating the proposed ViT-ReT transformer framework for human activity recognition using a contemporary human activity recognition dataset and comparing the results to state-of-the-art activity recognition models.
\end{enumerate}

The rest of the paper is organized as follows. Section \ref{sec:relatedresearch} discusses the general activity recognition model flow and current state-of-the-art activity recognition models. Section \ref{sec:background} gives background information regarding sequence-to-sequence models, transformers, and attention. Section \ref{sec:methodology} shows the application of the different TNNs to activity recognition models. Section \ref{sec:implementation} describes the practical implementation of the TNNs created for this paper. Section \ref{sec:experiments} describes the experiments performed on these models and Section \ref{sec:results} describes the experimental results. Finally, Section \ref{sec:conclusions} gives further discussion of these models and Section \ref{sec:futureresearch} provides future research directions.

\section{Related Research} \label{sec:relatedresearch}
\begin{figure}[tb]
    \centering
    \includegraphics[width=0.42\textwidth]{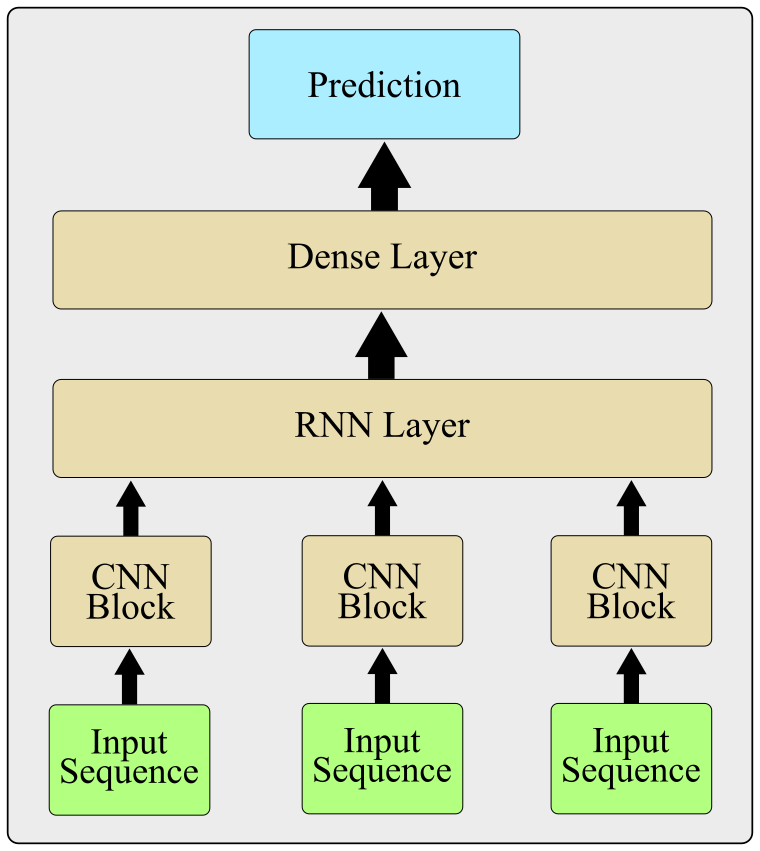}
    \caption{General activity recognition model flow.} 
    \label{fig:arflow}
\end{figure}

\begin{figure}[tb]
    \centering
    \includegraphics[width=0.495\textwidth]{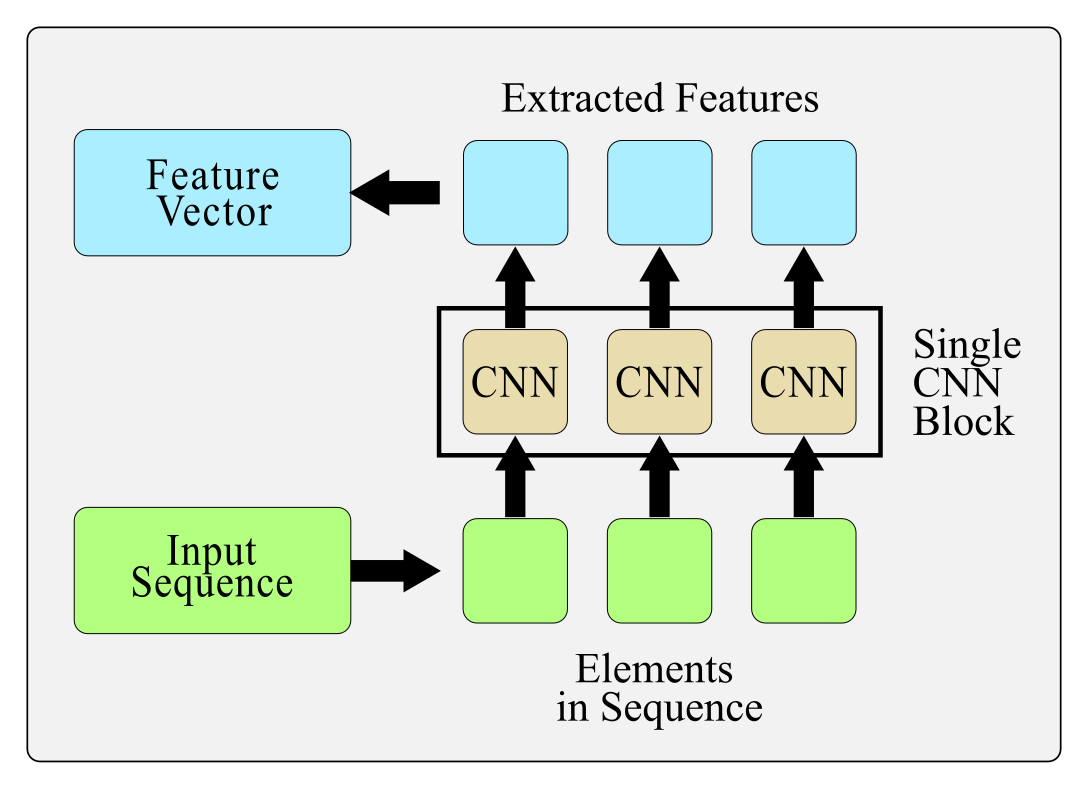}
    \caption{A single CNN block in the CNN layer of activity recognition model.}
    \label{fig:cnnblock}
\end{figure}

\begin{figure}[tb]
    \centering
    \includegraphics[width=0.495\textwidth]{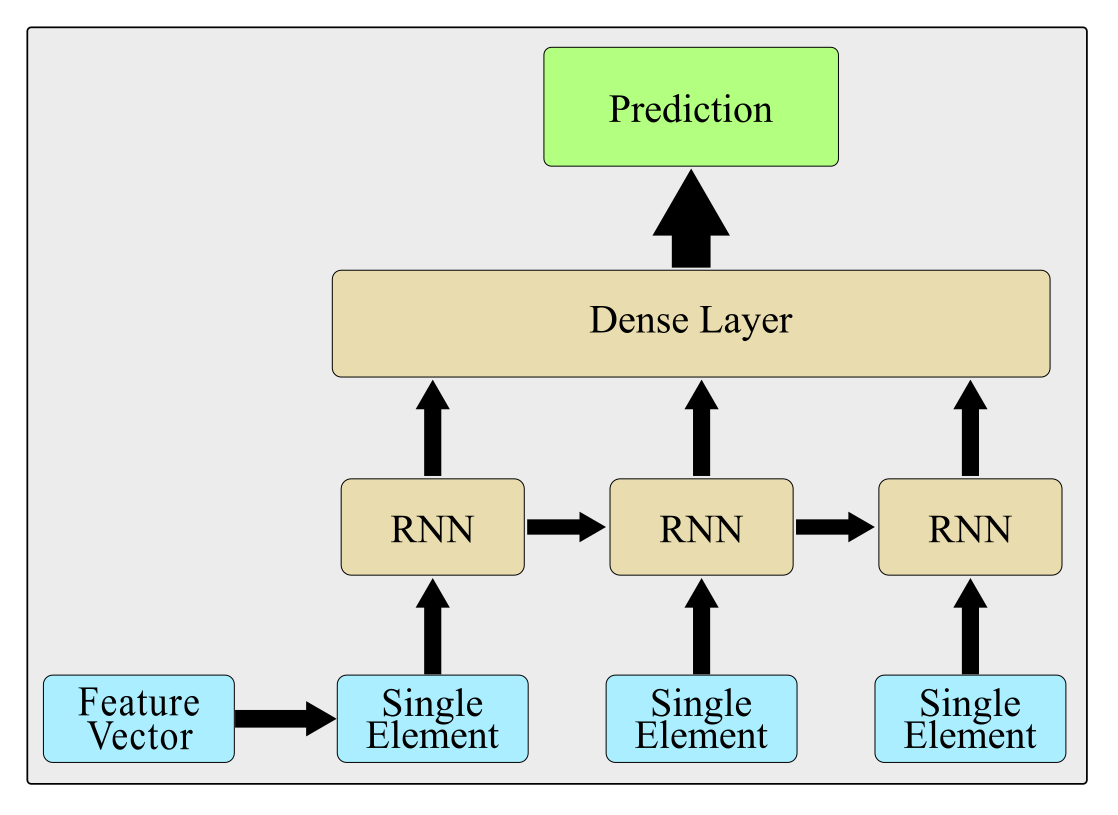}
    \caption{RNN and output layer of activity recognition model.} 
    \label{fig:rnndense}
\end{figure}

Activity recognition generally consists of two main phases: feature extraction on the input data sequence, and a combination of the extracted features into a time-dependent classifier. The standard version of this model uses a convolutional layer that is applied to each time step of the input. The input features extracted from this layer are kept in their original sequence and collectively input into a Recurrent layer (usually using LSTM units or GRUs), and then passed to a dense layer for classification \cite{YaoDeepSense}. This model structure is outlined in Fig. \ref{fig:arflow}, and Fig. \ref{fig:cnnblock} expands on the CNN layer, looking directly at a single CNN block applied to a single time step. Fig. \ref{fig:rnndense} expands on the RNN layers, showing the sequential nature of RNNs.\\
\indent The CNN layer usually contains multiple layers of convolutional filters mixed with pooling and activation layers. These layers can be very deep for improved performance. Similarly, the RNN layer can be made up of any number of RNN units, and there are usually at least 2 RNN layers stacked on top of each other, feeding into a dense layer that can also have any number of hidden layers and internal hidden units.\\
\indent While this model structure is very common in activity recognition, it has flaws that are inherent to the layers themselves. For convolutional layers, to get good and reliable classifications on large datasets with many classification categories, some form of deep and / or residual CNN is necessary \cite{HeDeepResL}. This causes complexity and speed issues on devices with limited capabilities, such as IoT or edge devices. The recurrent layers have similar problems, as the final RNN in the chain is dependent on all previous units in the layer. This means to calculate each output, all previous outputs must be calculated first, creating a bottleneck that can impact performance \cite{ViswaniAttention}, especially on edge devices close to the users.\\
\indent Sun et al. \cite{SunHAR} have used this model structure along with an extreme learning machine (ELM), a special type of dense neural network that requires a significant number of hidden nodes compared to general dense layers. They have used the OPPORTUNITY dataset \cite{OPPORTUNITY} to perform classification on 18 classes of gestures. Their ELM model is able to make accurate predictions on their gesture dataset, but like all general activity recognition models, it is not suitable for low resource environments or edge devices for real-time systems because of its complexity.
Russo et al. \cite{RussoExLow} have updated this model structure using only deep CNNs by first splitting their video data into three streams, red, green, and blue (RGB), optical flow, and slack foreground mask to perform classification on low resolution and extremely low-resolution images. While their approach removes the RNN layers and creates a simpler model as a result, it is done by using deeper and more complex CNNs. This results in a model that maintains the speed issues on limited capacity devices of the general activity recognition model. Attention has also been added to activity recognition models to improve accuracy. Ma et al. \cite{MaAttnSense} have applied attention layers after feature extraction from the CNN layers and again after time series classification from GRU RNN layers. While their model performs well over multiple datasets, it does not reduce the overall model complexity of activity recognition models. Therefore, their model is not suitable for resource-constrained environments.\\
\indent To alleviate the issues produced when using general activity recognition models in low resource environments, we propose the use of two types of TNNs, recurrent transformers and vision transformers, to replace both the CNN layers and the RNN layers in activity recognition models. This serves to reduce space and time complexity of the models, allowing them to run more efficiently on low resource devices.

\section{Background} \label{sec:background}
\subsection{Positional Encoding} \label{subsec:backgroundpositionalencoding}
When handling sequential data, it is important to make sure the information about the position of each element in the sequence is preserved. For example, if the initial sequence is the sentence “I had baked a cake”, it holds very different meaning to the sentences “I had a cake baked”, and the only difference is the position of a few elements in the sequence. For networks with recurrent connections, this information will inherently be conserved, but for networks without recurrences, that information must be preserved \cite{ViswaniAttention}.\\
\indent The first stage of a TNN is converting the input data into a tokenized and learned representation (or embedding). This representation will be a vector where each element in the vector is equivalent to a different concept that can be learned \cite{PatelEmbeddings}. For example, if the input sequence was a sequence of pixels, a three-element vector could be used, with an element for the red, green, and blue values for the pixel. Each element in the sequence is then converted to an embedded vector where each value represents the importance of each concept to the initial element. The different elements in the embedding vector can refer to any number of different concepts, and each value is learned during training. The size of this embedding vector is a hyperparameter that is problem dependent. For problems with many potentially important concepts, a larger embedding vector is used (the default for TNNs is 512 \cite{ViswaniAttention}).\\
\indent However, this embedding does not apply positional data to the input sequence. This is done by creating a positional encoding. This encoding will be the same length as the embedding vector for each element as it will be added directly to the embedding. This positional encoding will be a value that represents the importance of each concept in the embedding at each position in the initial sequence. This positional data can be learned during training, but for TNNs, a sinusoidal positional function is used as it showed similar performance to learned positional data \cite{ViswaniAttention}. The addition of this positional encoding to the element embedding allows the TNN to retain all positional information without any residual connections. In a TNN, each element is addressed individually with no connection to other elements in the sequence, so this positional encoding is a crucial step in the classification process to allow the relevant positional information to be maintained.
\subsection{Sequence-to-Sequence Models} \label{subsec:sequencetosequence}
\begin{figure}[tb]
    \centering
    \includegraphics[width=0.37\textwidth]{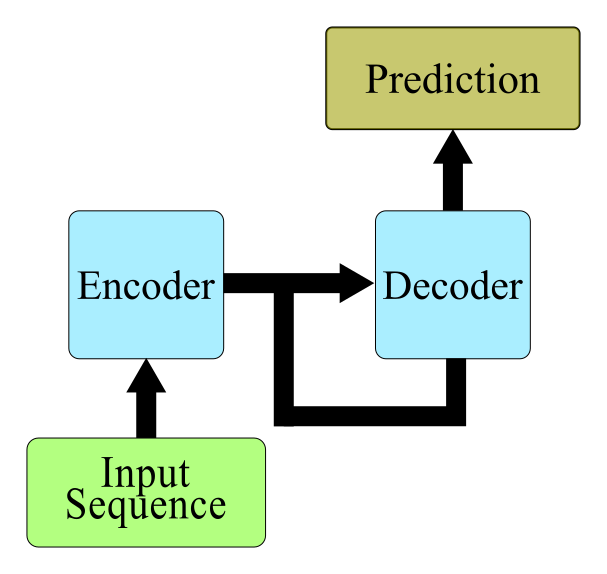}
    \caption{Overview of encoder-decoder structure.}
    \label{fig:seqtoseq}
\end{figure}

Sequence-to-sequence models are state-of-the-art neural networks that operate by learning to map an input sequence to an output sequence one step at a time \cite{SutskeverInfoProc}. Once trained, they are able to generate an output sequence from an input sequence, maintaining the meaning of the input through its own representation of the data that is used to inform the generation of the output sequence. This process was further expanded recently to create the encoder-decoder structure that is the basis for modern sequence-to-sequence models. The encoder-decoder model has two distinct parts that operate independently: an encoder and a decoder. The encoder takes an input sequence and encodes it into a vector that contains a representation of the meaning of the initial sequence. This encoding is then fed into the decoder which converts it into an output sequence in the target domain \cite{ChoRNNMT}. This general structure can be seen in Fig. \ref{fig:seqtoseq}. Some encoders (those built from RNNs) generate the internal encoding sequentially, beginning with the first element and proceeding one element a time \cite{ChoRNNMT}. Newer encoder architectures (specifically the Transformer architecture) can generate the entire input encoding in one step (for the transformer, this is done by using self-attention) \cite{ViswaniAttention}. The decoder takes the encoding, and at the first time-step generates a single element in the output sequence. The previously generated elements are used as inputs to the decoder at the next time step, which combines these elements with the encoding to generate the next element in the sequence. This process is repeated in the decoder until the entire sequence is generated.\\
\indent This type of model is most commonly associated with natural language processing, and most specifically machine translation. In machine translation, the input sequence is a sentence in some language that will be translated by the model. The encoder takes a sentence of positionally embedded words and creates an encoded representation of the sentence. This is either done all at once or one word at a time. This encoding is then given to the decoder, which uses it to generate the first word in the target language. It then uses the first word in the output sequence and combines it with the same encoding from the first step to generate the second word. Then the second word is used to generate the third, and so on. This process is repeated until the entire sentence has been generated in the target language. The two halves of the networks are trained together, and therefore complement each other in the learning process. However, the basic encoder-decoder structure, using RNNs for each, has issues as the length of the sequence gets longer \cite{BahdanauTranslation}, and it has trouble with interpreting and generating sentences longer than the sentences used to train it. This is accounted for and improved upon by the transformer model, which will be discussed in the next subsection.\\
\indent Since its inception, the encoder-decoder model structure has become the basis for the entire field of natural language processing. However, it is not immediately apparent how this type of model can be applied to the field of activity recognition. The direct application of sequence-to-sequence models to activity recognition will be describe in detail in Section \ref{sec:methodology}.
\subsection{Transformer Neural Networks} \label{subsec:transformerneuralnetworks}
\begin{figure}[t]
    \centering
    \includegraphics[width=0.35\textwidth]{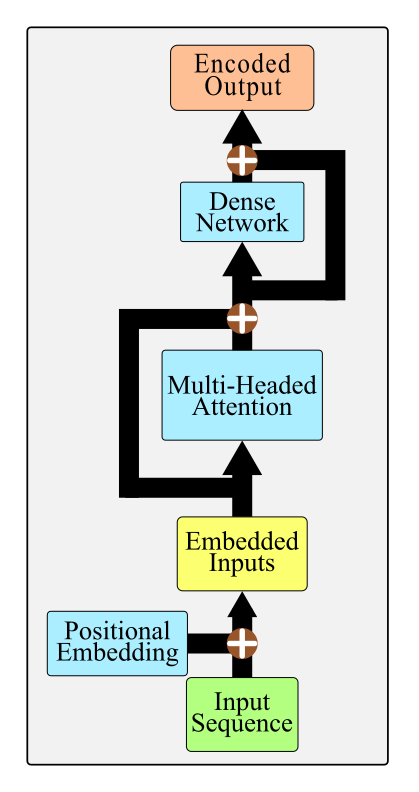}
    \caption{Transformer encoder.} 
    \label{fig:transformerencoder}
\end{figure}

TNNs are a special case of sequence-to-sequence model introduced in 2017 \cite{ViswaniAttention}. Viswani et al. have introduced a special type of self-attention called “scaled dot-product attention”. The particulars of this will be discussed in the next subsection. The self-attention of TNNs completely replaces the RNNs in both the encoder and decoder and allows the model to use the entire input sequence in the encoder at once. This alleviates one of the larger issues associated with RNNs, their inability to be used in parallel. The entire encoder can be run in parallel and all at once, which allow for large speedups of the network in both training and predicting. In TNNs, the decoder is still sequential and thus must be run one element at a time, but this eliminates half the bottleneck entirely. Models using TNNs, such as BERT \cite{DevlinBERT}, are extremely robust and have accuracies rivaling other sequential models.\\
\indent TNNs have a very similar flow to the general encoder-decoder, but with a few key differences. First, the input sequences are given a numerical representation and positional information is added to the input sequence to generate “embedded inputs”. This input is then passed into a self-attention layer that learns to “pay attention” to certain aspects of the input sequence. The attention the encoder uses is also said to be “multi-headed attention”, as the encoder uses multiple attention blocks. This is to allow each block to pay attention to different aspects of the input sequence and, ideally, preserve different levels of meaning at each. These are then passed into a feed-forward network that combines all the information learned in each attention head. After both the multi-head attention block and the feed-forward network, a residual connection is added to improve performance of the model, like the residual connections used in some deep CNNs like ResNet \cite{HeDeepResL}. The output from each section (i.e. multi-headed attention and dense network) is combined with the input, and the result is normalized. This allows the encoder to be more robust and allows it to learn better at each step. The encoder structure can be seen in Fig. \ref{fig:transformerencoder}.\\
\indent At this point, the encoded output sequences have been generated, containing a numerical representation for the meaning of the input sequence. This is then passed to the decoder. The decoder first passes all current and previously generated outputs to itself and applies multi-headed masked attention to them. This process will be described in the next subsection. The input is then fed into another multi-headed Attention block, but this time it is combined with the encoded output from the encoder block. The output of this layer is fed to a feed-forward network, like in the encoder block. There are again residual connections around both multi-headed attention blocks and the feed-forward network that are combined with the original input and normalized. Finally, the decoder output is passed to a linear layer with a softmax activation function to get a probability distribution as the output. This is then passed back to the beginning of the decoder as the input of the next layer. The decoder structure can be seen in Fig. \ref{fig:transformerdecoder}, and the overall structure of the Transformer can be seen in Fig. \ref{fig:transformer}.\\
\indent TNNs have quickly become the most important neural network for sequence-to-sequence tasks. While it is clear it is extremely powerful, it is still not obvious how this can be applied to video data or activity recognition as a whole. This process will be described in detail in the Section \ref{sec:methodology}. 
\subsection{Self-Attention} \label{subsec:selfattention}
Self-attention is a process by which a neural network learns which parts of the input it should focus on. While there are many kinds of attention, such as additive attention \cite{BahdanauTranslation}, this section will focus on scaled dot-product attention \cite{ViswaniAttention} as it is the basis for the transformer models created by Viswani, et al. \cite{ViswaniAttention}. Attention is derived from database queries. The process begins with a set of key-value pairs and involves mapping a set of queries to those keys and combining the result with the initial values to get the output. When given a set of queries Q, keys K, and values V, where Q, K, and V are all matrices, and dk is the number of dimensions of the keys matrix, the scaled dot-product attention can be calculated by eq. (\ref{eq:attention}) [14, eq. (1)]: 
\begin{figure}[tb]
    \centering
    \includegraphics[width=0.35\textwidth]{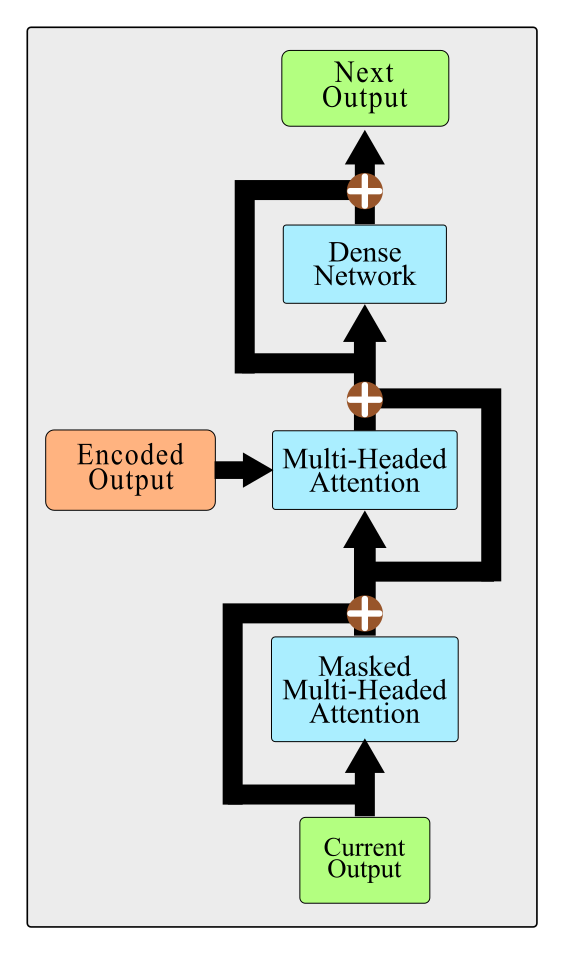}
    \caption{Transformer decoder.} 
    \label{fig:transformerdecoder}
\end{figure}
\begin{figure}[tb]
    \centering
    \includegraphics[width=0.495\textwidth]{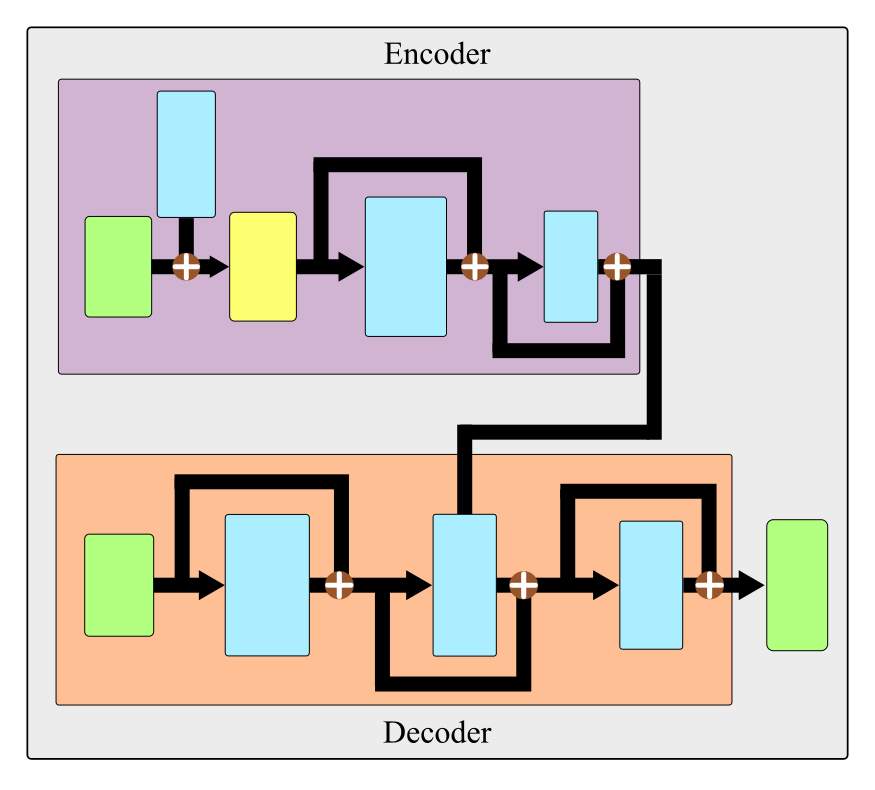}
    \caption{Transformer.} 
    \label{fig:transformer}
\end{figure}
\begin{equation} \label{eq:attention}
Attention(Q,K,V) = softmax\left(\frac{QK^T}{\sqrt{d_k}}\right)V
\end{equation}
\indent While not immediately obvious, this equation is based on the equation for cosine similarity \cite{ReTEx}. Looking at that equation will help to better explain what Attention is doing. The cosine similarity of two vectors A and B is given by eq. (\ref{eq:cosinesimilarity}):
\begin{equation} \label{eq:cosinesimilarity}
CosineSimilarity(A,B) = \frac{A \cdot B}{\rvert \rvert A \lvert \lvert \: \rvert \rvert B \lvert \lvert}
\end{equation} 
\indent This equation returns a value between +1 and -1, describing how similar vectors A and B are, which is the equivalent of the cosine of the angle between the two vectors. First the dot product between the two is computed, then it is scaled by the product of the magnitude of A and B. This is similar to a portion of the original equation shown in eq. (\ref{eq:inner}).
\begin{equation} \label{eq:inner}
\frac{QK^T}{\sqrt{d_k}}
\end{equation} 
\indent In this case, the Queries matrix and the transpose of the Keys matrix serve as A and B, and instead of scaling by the product of their magnitudes, the result is scaled by the root of the number of dimensions of the keys matrix. This scaling value was chosen as an extension to dot-product attention to give it better performance as the keys matrix becomes larger. This was done because the dot product at higher dimensionalities of the keys matrix will be very large and would have a negative effect on the performance of the softmax function otherwise \cite{ViswaniAttention}. Using matrices for A and B instead of individual vectors allows for the computation of multiple dot products at the same time. The result of the scaled dot-product is then passed through a softmax layer, giving us a probability matrix that is used as the weights for the values matrix (when performing masked attention, the mask is also applied before the softmax layer). This process can be seen visually in Fig. \ref{fig:dotproductattention}. This can also be written to resemble typical neurons more closely, as in eq. (\ref{eq:typcialneurons}):
\begin{equation}\label{eq:typcialneurons}
y_k = Wx_k , 
\end{equation}
where
\[ W = softmax\left(\frac{QK^T}{\sqrt{d_k}}\right)\]
and
\[
x_k = V
\]
\indent The encoder and decoder attention blocks each have different inputs for the queries, keys, and values. For the encoder, the queries, keys, and values matrices are 3 identical copies of the input sequence. This seems counter intuitive that all 3 would be the same matrix, but this is a crucial step in obtaining the meaning of an input sequence. For example, take the input sentence “They chose to eat pizza”. Each row element in the keys matrix corresponds to a word in the original sentence, and the data for each row element corresponds to the vectorized and learned embedding of that word. When this is transposed and the scaled dot-product is calculated, the result will be a 5-by-5 attention matrix where each row or column element is still equivalent to the corresponding word in the initial sentence, but the data no longer represents a single word. Instead, the data at column j row i now corresponds to how similar word i is to word j in the original sentence. Once the softmax is applied to the attention matrix, it will give the similarity of i and j as a value between 0 and 1, with 1 meaning they are completely alike, and 0 meaning they have nothing in common. In the final attention matrix, if row 5 column 4 has a value of 0.8, that means the meaning of the fourth and fifth words (in our example, “eat” and “pizza”) are heavily tied together. If instead row 3 column 1 has a score of 0.1 (“they” and “to” in our example), the meanings of the two words are not strongly related in this sentence. This process as well an example softmax attention matrix is shown in Fig. \ref{fig:attentionmatrix}.\\
\begin{figure}[tb]
    \centering
    \includegraphics[width=0.25\textwidth]{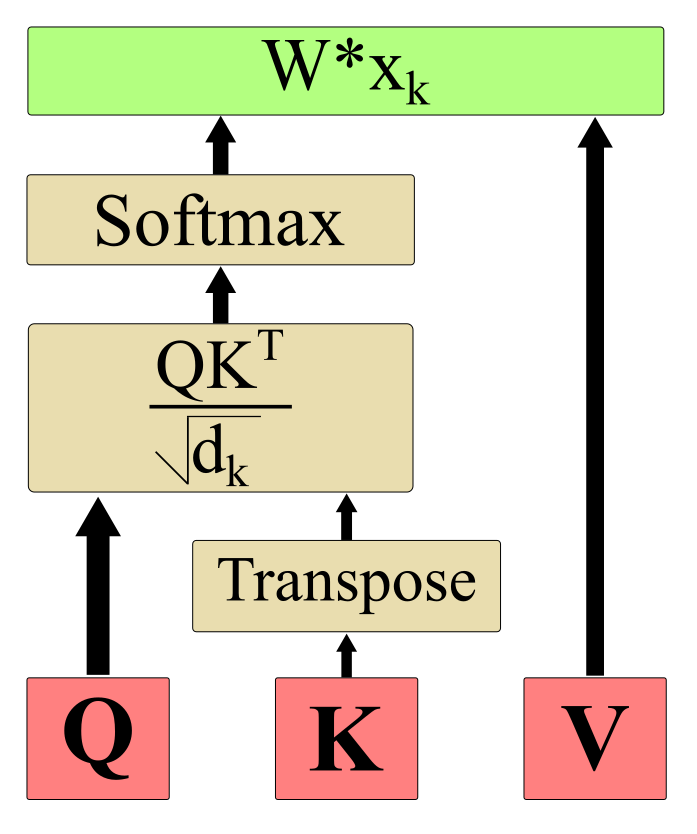}
    \caption{Scaled dot-product attention.} 
    \label{fig:dotproductattention}
\end{figure}
\indent This final softmax attention matrix is then used to weight the values to give the output of the attention block. Combining the values with the weight’s matrix will create a final matrix where each row element still corresponds to the initial words in the input sentences. However, the data of each row element will now be a weighted dot product that combines how much each word in the sentences impacts the word at the current row index. The output of this layer is demonstrated in Fig. \ref{fig:matrixexample}.\\
\indent The decoder attention blocks operate the same way but with 2 key distinctions. First, for the masked attention block, the queries, keys, and values are the current output sequence that has been generated (or in the case of training, the correct output sentence). The only difference for the decoder block is that the attention matrix is masked before the softmax function is applied. The mask simply sets all values in the sequence beyond the last generated output element equal to 0. This is done because, during training, the correct output sequence is fed into the decoder instead of its own output which will be the input for the decoder in practice. This ensures correct learning, even after incorrectly predicting the earlier elements of the sequence. At each step, even if the decoder previously made an incorrect prediction, it will have the correct and necessary information to predict the next element in the sequence. Therefore, the mask is used by the decoder to hide correct information it should not see yet (i.e. elements in the sequence beyond what it has already generated). This ensures it only makes predictions based on the information it would have in practice, even though it has the entire output sequence during training. An example matrix showing the masking process can be seen in Fig. \ref{fig:mask}.\\ 
\begin{figure}[tb]
    \centering
    \includegraphics[width=0.495\textwidth]{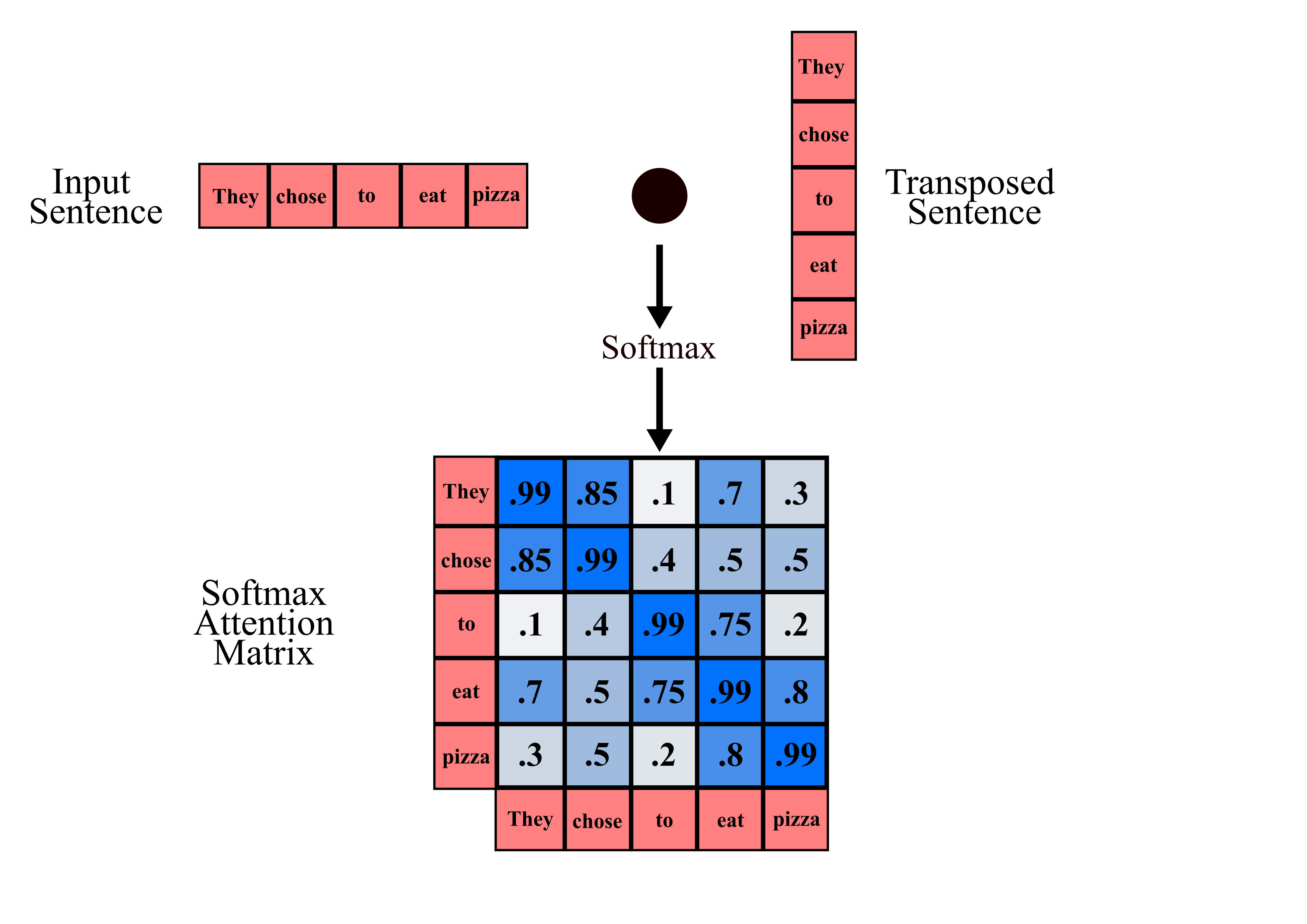}
    \caption{The process and results of scaled dot-product attention.} 
    \label{fig:attentionmatrix}
\end{figure}
\begin{figure}[tb]
    \centering
    \includegraphics[width=0.33\textwidth]{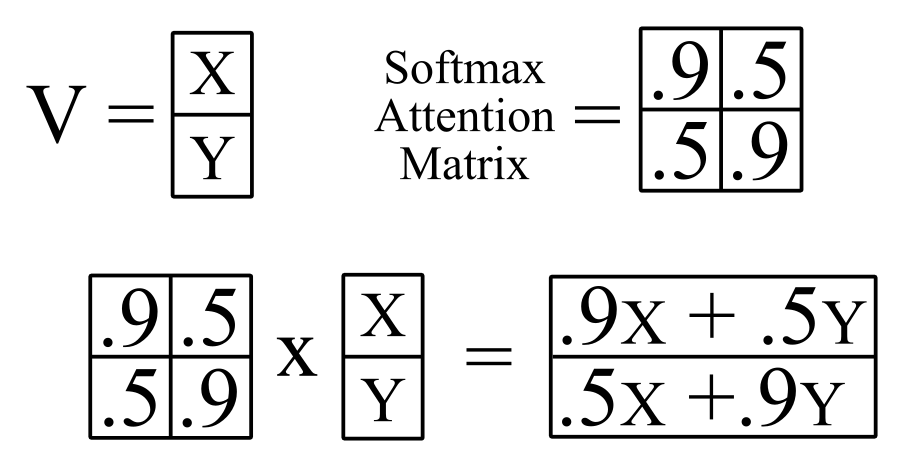}
    \caption{Example weighted values matrix.} 
    \label{fig:matrixexample}
\end{figure}
\indent Finally, the result of this masked attention block will be a matrix containing the relevance of each word in the currently generated output sequence to every other word in the sequence. This relevance matrix will contain all contextual information that is currently known about the output sequence. This matrix is used as the values for the second attention block in the decoder, and the encoded output of the encoder block is used as the queries and keys. The encoded output still maintains all contextual information from the initial input. Using the encoded output as the queries and keys allows us to calculate a set of weights directly from the encoder output, and when those weights are applied to the relevance matrix from the first attention block, it will apply the encoded contextual information of the input sequence to all currently known contextual information about the output sequence. This gives us a contextual matrix that contains all known context from the input sequence and current output sequences. This is then combined with a residual connection, normalized, and passed through one last feed-forward network, normalized one last time, and passed into a linear classification layer with softmax activation function to predict the next element in the output sequence. These final layers all apply extra preprocessing to their inputs to help improve consistency of the output predictions.
\begin{figure}[tb]
    \centering
    \includegraphics[width=0.495\textwidth]{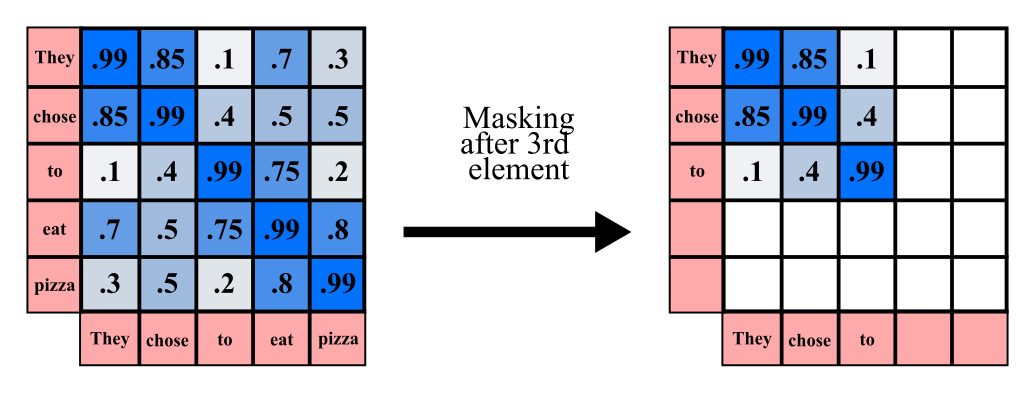}
    \caption{Masking of a Softmax attention matrix after the third element.} 
    \label{fig:mask}
\end{figure} 
\subsection{Multi-Headed Attention} \label{subsec:multiheadedattention}
The attention used by TNNs is said to be “multi-headed”. This is because it uses multiple attention blocks in parallel. Before being passed to any of the attention blocks, the queries, keys, and values are all linearly projected using a different learned linear projection for each attention head. This allows each head to focus on different sections of the input data. Each head then performs attention normally. The output of each attention block is then concatenated and linearly projected to the desired output shape. This process allows for more accurate classifications by allowing the model to pay attention to multiple different aspects of the input sequence.\\
\indent For a more detailed description of both TNNs and scaled dot-product attention, see Viswani et al., \cite{ViswaniAttention}. Much like the TNNs, scaled dot-product attention is the current state-of-the-art attention method that is applied to many domains, and is even used beyond TNNs \cite{DebnathBodyPose}. However, the direct application of TNNs to activity recognition is still not obvious.

\section{Methodology} \label{sec:methodology}
\subsection{RNN Layers} \label{subsec:rnnlayers}
The application of TNNs for the RNN layers of the generalized activity recognition model seen in Fig. 1 and Fig. 3 is the simplest of the two applications. While TNNs were designed for, and most used for, machine translation tasks, there is nothing inherent to the model requiring word embeddings to be used as the inputs to the model. The encoder can take any sequential input, so long as the relevant positional information has been encoded in the sequence. It will extrapolate out the relationships between all elements in the sequence. So, if instead of a sequence of words, the input was a sequence of frames, where each frame was embedded with the information relevant to its order in the video, the encoder would be able to find the relationship between all frames in the video and use this to create an output containing the “meaning” of the video. While the application of the encoder to get meaning from the frames means video data can be used as the input to the encoder, an important question arises: What sequence will be generated by the decoder?  Shockingly, the answer is none.\\
\begin{figure*}[t]
    \centering
    \includegraphics[width=\textwidth]{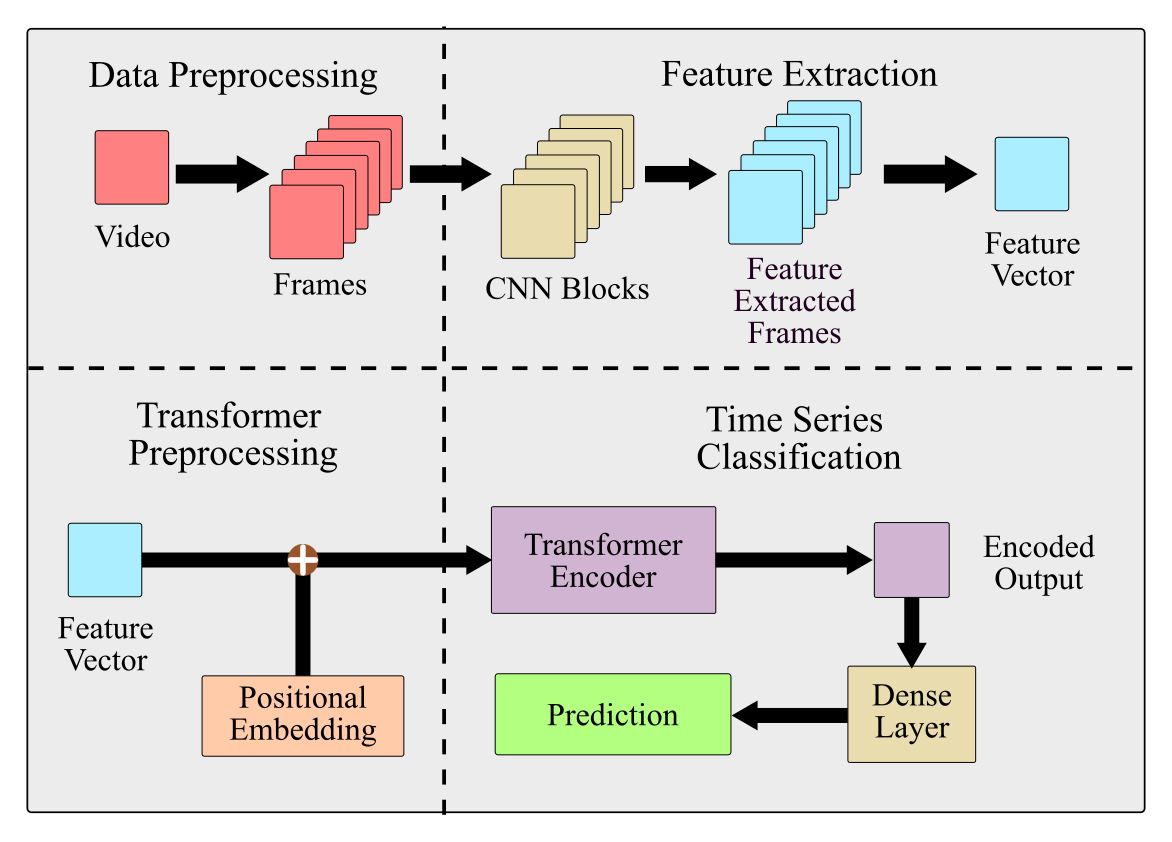}
    \caption{Overview of the activity recognition pipeline using the transformer for time series classification instead of the typical RNN.} 
    \label{fig:transformerarflow}
\end{figure*}
\begin{figure}[ht]
    \centering
    \includegraphics[width=0.37\textwidth]{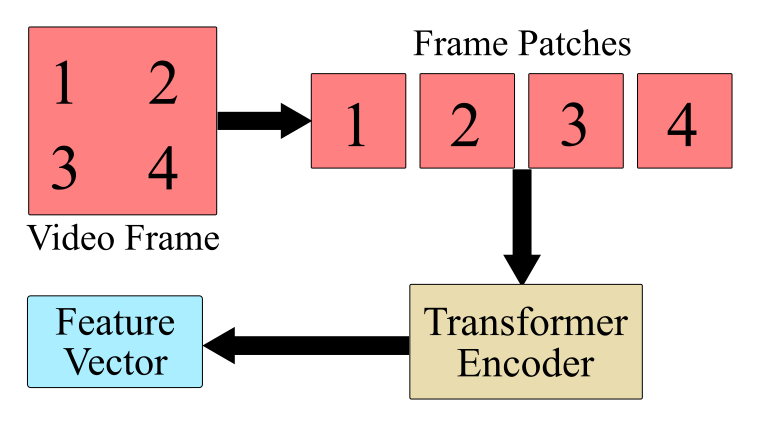}
    \caption{Splitting a frame into patches and passing them to a transformer for classification} 
    \label{fig:patching}
\end{figure} 
\indent In the case of video data (or any time series data that does not generate an output sequence), there is no need to use the decoder at all. Instead, the output of the encoder is fed directly into a flattening layer to make the data 1-dimensional, and then to a Dense layer to make predictions. Conceptually, this structure makes sense. After the multi-headed attention layer, the Encoder has created a matrix from the initial sequence that has been weighted by what the model has determined is important about each element. This is passed into a Feed-Forward layer to create an encoded output, which can be thought of as an encoded representation of the “meaning” of the input sequence. So instead of using said “meaning” to generate a sequence, this “meaning” can instead be used to classify the input, which is exactly what our proposed model does. Further, because our proposed model does not use the decoder, there is no bottleneck in this section. Unlike when RNNs are used to generate the final representation of the data, there is no need to wait on the completion of any chains of values waiting on previous states. Even in the application of the TNN to sequence-to-sequence models, the data gets bottlenecked at the decoder step. Since all values needed to compute the output of the encoder are present at the start of computation it is easily parallelizable, which allows for increased computation speed. This model structure can be seen in Fig. \ref{fig:transformerarflow}.\\
\indent While the application of TNNs to any other time series data would follow similarly, it is not easily apparent how to apply this to non-time dependent data, such as single images, in a way that it could replace CNNs.
\subsection{CNN Layers} \label{cnnlayers} 
TNNs needs to be applied to data in a sequence. However, the TNN itself does not learn anything from the positional data of the input sequence. This is all done externally by the positional embedding of the input sequence data. So, if the non-time series data was somehow made to be sequential, with the information about its locality still applied by the positional embedding, then the data could be used in the same manner as any time series data. It is by this process that the TNN is applied to the individual frames of a video, or to any other non-sequential data. To turn an image into sequence data for a TNN, it is first split into patches, or smaller subsections of the original image. These patches are then grouped in sequence to be used as the input to the TNN. This process can be seen in Fig. \ref{fig:patching}. The size of these patches is a problem dependent hyper-parameter similar to CNN filter size. For images with large object scales, larger patches may be used, but for images with much smaller scale, small patches are necessary.\\
\indent The TNN will then perform a pseudo feature extraction, by determining the internal “meaning” of the image, and using a dense layer to classify this output, in the same manner as for time series data. The patching of an image and using it as a sequential input to a TNN is the process taken by a vision transformer \cite{Dosovitskiy16Words}. Vision transformers have been shown, when trained on a sufficiently large dataset, to perform as well as state-of-the-art deep residual CNN networks \cite{Dosovitskiy16Words}, while remaining more lightweight and efficient than long CNN chains. The last important step in the application of TNNs for activity recognition is utilizing the vision transformer alongside the recurrent transformer.
\subsection{RNN and CNN Layers} \label{rnnandcnnlayers}
An important note before this section is it is not necessary to implement TNN models for both halves of the activity recognition model.  The recurrent transformer (ReT) can replace the RNN layers only and train on the features extracted by a deep residual CNN more efficiently and with as much accuracy as the typical LSTM or GRU chains. Alternatively, the vision transformer (ViT) can be used to replace the deep residual CNN to perform feature extraction from the frames and pass the data to the RNN layers. Still, combining the ReT and ViT will prove to alleviate the largest amount of complexity and efficiency issues encountered by current activity recognition models.\\
\indent To use a ViT, it should first be trained separately on a large image recognition dataset, such as ImageNet \cite{DengImagenet}. The ViT needs a large dataset to train, and there are no sufficiently large activity recognition datasets that would allow the ViT to achieve adequate performance if first trained in conjunction with a ReT. Once trained, it will be transferred to be the backbone of the feature extraction section of the new activity recognition model. Then, for a specific video input, a copy of the trained ViT will be applied to each frame. The output from this layer will be a series of encoded image data generated from the patches of each frame in the sequence. The encoded image data will be in the same order as it was in the initial video, and it is this sequence that will be classified by the Transformer. In the next section, we elaborate our implementation of both a ReT model and a ViT model.

\section{Implementation} \label{sec:implementation}
We have built our model in Python 3.8.4 using TensorFlow through the Keras API, utilizing Numpy for data manipulation and OpenCV for video processing. The implementation of our model comprises of three main stages: (i) data preprocessing, (ii) custom layers, and (iii) model controllers. This section discusses the key components of our implementations. Afterwards, the important hyperparameters for both ReT and ViT models will be discussed, followed by an explanation of the ViT used.
\subsection{Data Preprocessing} \label{subsec:datapreprocessing}
The first step in training a ViT is preprocessing the video files in the datasets to be ready to train the model. This is done by subsampling a selected number of frames from the videos to create consistent length sequences. The entire video is not used for a few reasons. It is unnecessary, as most frames contain the same data as their surrounding frames and using an entire video would significantly increase model complexity with little to no tangible benefit to accuracy. This also allows for the use of any length of video (even if a video has more frames than the number used for the ViT) since it will be shortened to the correct sequence length for the model. The selected frames are then resized to 224x224 pixels, which is the default size for ResNet50 \cite{HeDeepResL}. The ViT has a default image size of 384x384 \cite{Dosovitskiy16Words}, but it can be used for images of any size so only 224x224 images were used for testing to ensure the same testing conditions for both models. After resizing all selected frames, they are appended to an array containing all processed video files, and the associated training label is appended to the same index in a labels array for later training and testing. These two arrays, along with an array containing the file path to each video file are saved for later use. This process can be seen in Fig. 14.\\
\indent Instead of completing these preprocessing tasks before training, they could instead be applied as layers at the beginning of the model, but we choose not to do this. This is done to save time when training models, as the data processing would need to be completed each time the models are trained, and for the dataset used this takes a lot of time (sometimes over 40 minutes on its own). This also allows for more accurate data collections from the models as each layer is then able to be monitored individually instead of only being able to watch the entire model at once.
\subsection{Custom Layer} \label{subsec:customlayers} 
To implement the ReT and ViT, three custom layers were created: PositionalEncoding, Encoder, and Patches. A functional wrapper, the BuildEncoder function, is also created to make creation of an encoder more seamless. Finally, a data generator is created to facilitate the use of a large dataset.\\
\subsubsection{PositionalEncoding} \label{subsubsec:positionalencodinglayer} 
This layer applys the positional data to the input sequence. It takes as an input the length of the sequence and the size of the linear project used on the input. The length of the sequence will either be the number of frames extracted from each video, or the number of patches extracted from an image, depending on whether the ReT or ViT is being used. The layer creates a positional encoding space and adds the associated positional data to each element. It also linearly projects the input to any size chosen through the linear layer (usually the size is left alone or slightly downsampled, but any could be selected).
\begin{figure}[tb]
    \centering
    \includegraphics[width=0.5\textwidth]{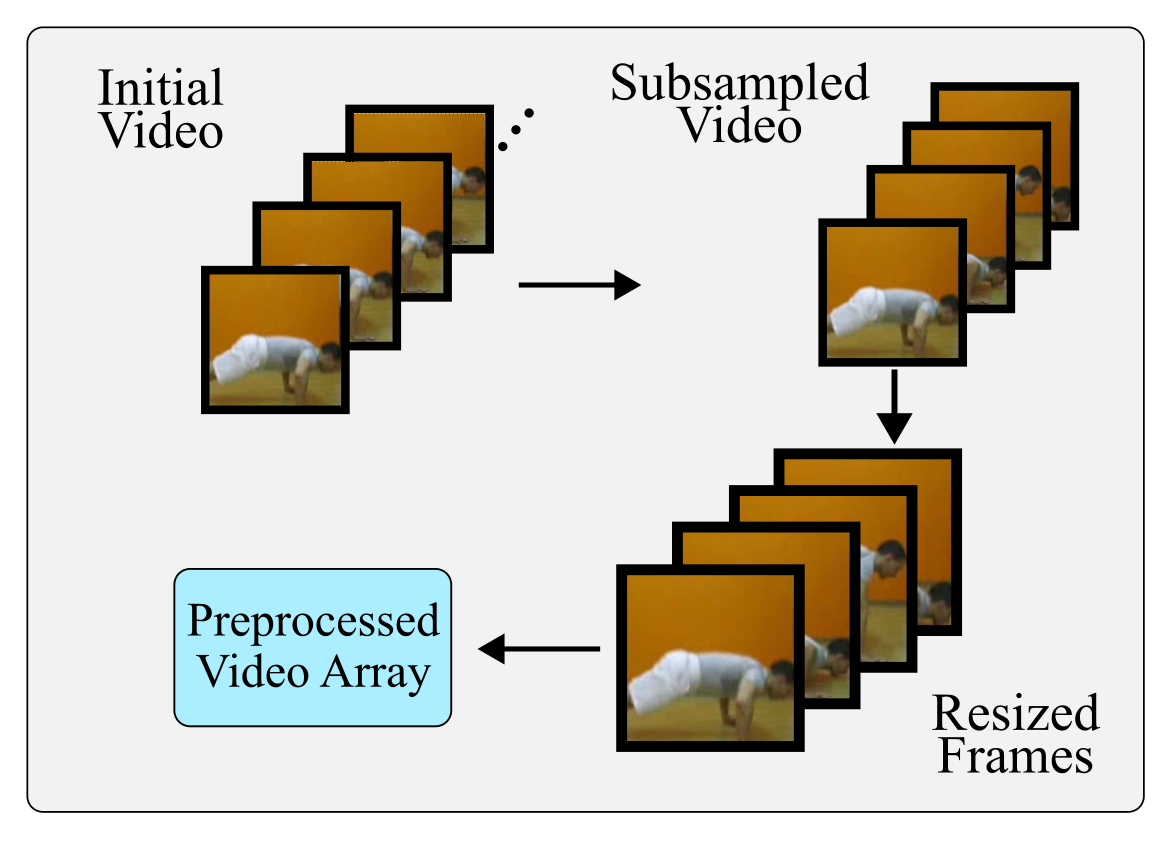}
    \caption{Video preprocessing steps.}
    \label{fig:preprocessing}
\end{figure}
\subsubsection{Encoder} \label{subsubsec:encoder} 
This layer serves as the transformer. Since none of the data generates a sequential output, the decoder is not implemented. The layer first takes as input the size of the input (which is the same as the size of the linear projection from the PositionalEncoding layer). The input shape will be conserved through the encoder as it is uses the size of the final dimension of the input vector as the shape of the final dense layer in the encoder. The next input is the shape of the internal dense layer. This can be any size to allow for more hidden neurons, but is usually set to twice the input size. The next input is the number of heads for the multi-headed attention in the encoder, usually 6 but could be any value. Finally, the activation function for the internal dense layer can be passed but is by default an ReLU activation function. This layer first computes the attention matrix from the given inputs using the built in MultiHeadAttention function from Keras. The output of this is the weighted values matrix. This is then normalized and passed into the dense layer, and normalized again, at which point we have the predictions or feature extracted vector depending on which TNN is being trained.\\
\subsubsection{Patches} \label{subsubsec:patches}
This layer splits an image into patches. It is given an input of the length of one side of the patch in pixels (8 was used but could be any value, smaller patch sizes will create more patches). This layer uses TensorFlow’s built in patch extraction function to create patches, then reshapes the resulting tensor to maintain the correct shape of the output. If the input is a (Batch size, 64, 64, 3) shaped image tensor and patches are 8x8, it will return a tensor with shape (Batch size, 64, 192) containg all 64 created patches from the image [192 = 8x8x3]. The patches are not kept in a (8,8,3) tensor, but instead the pixel data from each pixel in the patch is placed into a 1-Dimensional tensor. This is done to ellimnate future reshaping of each individual patch.\\
\subsubsection{BuildEncoder} \label{subsubsec:buildencoder}
This is a functional wrapper used to build a specific TNN layer according to user specifications. It takes as input the sequence length (number of frames) and the size of the final dimension of the data tensor. It also takes as input all relevant information for the PositionalEncoding layer (embedding layer shape) and Encoder layer (dense layer shape and number of attention heads). It then creates and returns a Keras Model Transformer according to this specification. To use this model, it will then be passed an input tensor with the expected shape, which will be passed to the positional encoding. After the positional data is applied, it will be passed to the Encoder where it performs attention and creates a weighted values matrix which is normalized, passed through a dense layer, and a softmax layer to receive a classification from the model.\\
\subsubsection{Data Generator} \label{subsubsec:datagenerator}
A generator is  used to train and test the models on different datasets. After initialization, it is called instead of accessing the dataset directly. When called, it returns the batch from the datset at the current index. This allows the system to only store the current batch of data in RAM instead of the entire dataset, making more efficient use of the available resources and allowing for larger datasets to be used. Without a generator, the entire dataset must be loaded into RAM, as well as all parameters for the model that is training. This puts an unnecessary burden on the resources of the computer and does not allow for large datasets. However, there would be an increase in training and testing speed if the dataset could be loaded into RAM in its entirety as with a generator some time is required to access each subsequent batch of data.\\
\subsection{Model Controllers} \label{subsec:modelcontrollers} 
The model controllers are files that are used to either create the models, run them, test them, or output data from them. Any parameters that are not mentioned here have been mentioned previously and will be covered again in the Hyperparameters subsection.\\   
\subsubsection{Create Models} \label{subsubsec:createmodels}
The model creation file has 5 functions, each of which creates a different model. The ResNet\_Model function generates an instance of the pretrained ResNet50 with ImageNet weights and global average pooling. This is applied over an input sequence using the Keras TimeDistributed wrapper, and is used to perform general feature extraction. The LSTM\_Model function generates a specified number of LSTM layers following the general activity recognition RNN layer architecture. All layers will have the same specified number of LSTM units. The Vision\_Transformer\_Model function creates an instance of a pretrained Vision Transformer model. The Untrained\_VisionTransformer\_Model function creates a model ready to train for image recognition. It will have a specified number of transformers stacked on top of each other. Finally the Transformer\_Model function creates a Transformer model with the specified number of transformer layers for classification of video data.\\
\subsubsection{Run Model} \label{subsubsec:runmodel}
The model running file contains three functions that are used to train, test, and evaluate a model, and to make predictions. First, the FitModel function is passed a model and the data to train on, as well as the batch size and the number of epochs to use. It trains the given model with the data provided, with the specified batch size and number of training epochs. The PlotWholeModel function is a helper function that visualizes a model as well as all functional layers. When a functional layer is found, the function calls itself, passing the functional layer instead of the whole model to dispaly the model hidden in the functional layer. PredictModel is used to make predictions on a model without training by passing it a model (which is assumed to be pretrained) and a features dataset to make predictions on.\\
\subsubsection{Test Models} \label{subsubsec:testmodels}
Testing of the models is done using 2 main files: a file that writes a test configuration JSON object (tests.json) and a file that reads the object and runs tests acording to those specifcations. The object contains a list of setup values and a tests to perform, indexed by the model that is being tested and the specific attribute of that model that is being tested. A list of default values for each important parameter is also included to allow for easy changing of the defaults in subsequent testings. This object is read by a TestModel file which loops over all the test types included in the configuration object file and uses the default and setup values alongside the specific model configuration being tested to create a new model and test it. The results of these tests are output and written to files for later interpretation.\\
\subsubsection{Output Model Data} \label{subsubsec:outputmodeldata}
The output model data file contains a function that creates progress bars over the command line for better visual represenatation of the progress completed by tasks. It also contains a class definition with 13 functions. The class creates an object that is capabale of storing timing and memory data before and after any event by calling an associated method for the object. This allows all other files to be much cleaner as they do not need to manually track their own data for output. It is also capable of writing all stored data to the command line, to a text file, and to a csv file. This allows for easy recording of test data for later interpretation.
\subsection{Hyperparameters} \label{subsec:hyperparamters}
This section contains a listing of each hyperparameter for the model, and a brief description of each, including the values used when training the model.
\begin{enumerate}
    \item \textit{SEQUENCE\_LENGTH}: The number of frames to keep from each video (usually 20). 
    \item \textit{IMAGE\_HEIGHT}: The height in pixels to resize the video frames for activity recognition (usually 224).
    \item \textit{IMAGE\_WIDTH}: The width in pixels to resize the video frames for activity recognition (usually 224). 
    \item \textit{projection\_dim}: The size of the linear projection of the input sequence by the PositionalEncoder (usually left at or as close to input size as possible) 
    \item \textit{dense\_dim}: The number of neurons in the TNN’s feed-forward network’s hidden layer (this does not change the output shape as it is later projected back to the input shape; it is normally 2x input shape) 
    \item \textit{activation}: The activation function for Transformer’s Feed-Forward Network’s hidden layer (defalts to ReLU) 
    \item \textit{num\_heads}: The number of heads for multi-headed attention.
    \item \textit{patch\_size}: The length of one side of the patch to create in pixels (usually 8 [creates 8x8 patches from image]).
    \item \textit{LSTM\_layers}: The number of LSTM layers in an LSTM model.
    \item \textit{LSTM\_units}: The number of LSTM units in each layer of an LSTM model.
    \item \textit{transformer\_layers}: The number of transformer layers for a recurrent transformer or vision transformer model.
    \item \textit{cateogries}: The number of video or image categories to train on (determined by the dataset being used) 
    \item \textit{batch\_size}: The number of examples to train on before updating the weights of the model.
    \item \textit{epochs}: The number of times to train the model using the entire training dataset.
\end{enumerate}
\subsection{Vision Transformer} \label{subsec:visiontransformer}
\begin{figure}[ht]
    \centering
    \includegraphics[width=0.33\textwidth]{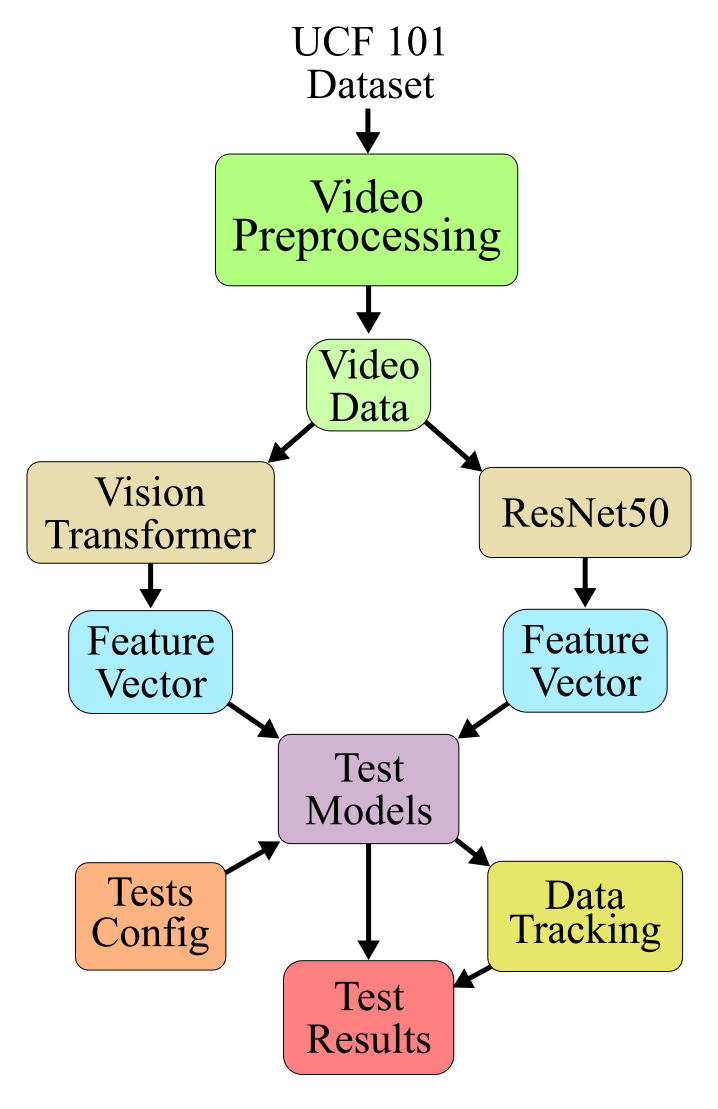}
    \caption{Basic dataflow of model testing.} 
    \label{fig:dataflow}
\end{figure}

To allow for quicker training and testing of models for activity recognition, we have used a pretrained implementation of the vision transformer \cite{ViTKeras}. This implementation utilizes the vision transformer weights pretrained on ImageNet by Steiner et al. \cite{TrainViT}. We have selected these pretrained weights to ensure that both ResNet and the vision transformers are trained on the same data in the same manner as Dosovitskiy et al. \cite{Dosovitskiy16Words} when they initially created the vision transformer. These model implementations are then used for activity recognition in tandem with the other model implementations as outlined in this section. The interaction between all files in this implementation can be seen in Fig. \ref{fig:dataflow}.

\section{Experiments} \label{sec:experiments} 
To compare the typical activity recognition model to the transformer activity recognition model, we have built multiple different models and have trained on various subsets of the UCF 101 dataset, which is one of the largest and most challenging activity recognition datasets. The UCF 101 dataset is a set of 13,320 human activity videos, each belonging to one of 101 different categories \cite{UCF101}. Each category contains approximately 130 videos, each approximately 10 seconds in length. The categories can be divided into 5 action types: Human-Object Interaction, Body-Motion Only, Human-Human Interaction, Playing Musical Instruments, and Sports.\\
\indent In this work, we have used a subset of 4, 20, and 50 action categories alongside the entire 101 action category dataset to show the models’ performance and scalability over datasets of different sizes. The subset with 4 categories has 478 videos, the subset with 20 categories has 2500 videos, the subset with 50 categories has 6,567 videos, and the total dataset has 13,320 videos. Models are trained and optimized on each data subset. The categorical cross entropy loss and accuracy of the training are tracked and will be compared in the results section (Section VI). The video files are subsampled to include only 20 total frames from each video to standardize the data and reduce overall model complexity. The frames are then resized to 224x224, the standard image size for ResNet50.\\
\indent The models are trained using both ResNet50 and Vision Transformer as feature extractors. Both feature extractors are pretrained using ImageNet 2012. Direct comparisons of ResNet50 and the Vision Transformer are not performed as extensive testing of these two models using ImageNet 2012 was performed by Dosovitskiy et al. \cite{Dosovitskiy16Words}.  ResNet50 and the Vision Transformer are trained using the same input sequence length and video frame size. \\
\indent The LSTM and Transformer models are trained and optimized according to their model parameters. For the LSTM model, the number of LSTM layers and the number of LSTM units per layer are changed, while for the Transformer, the number of Transformer Layers, the size of the feature embedding, the number of neurons in the internal dense layer, and the number of attention heads are all changed. These values are initially assigned using default values, but the parameter value that optimizes loss for the current data subset is then used for subsequent tests within that subset. For example, the LSTM model was tested with 1, 2, 4, and 6 layers. If using 2 layers resulted in the smallest loss while training the 4-category data subset, then when testing the optimal number of LSTM Units in the 4-category data subset, 2 LSTM layers would be used. After testing all model parameters, the values are then set back to the original default values for the next testing set.\\
\indent A breakdown of all the tests performed can be seen in Fig. 16. It shows a list of all tests performed for both the LSTM and Transformer models, as well as showing that they are repeated 8 times each. The tests for each model can be seen explicitly below:\\
\begin{enumerate}
    \item LSTM \begin{enumerate}
        \item Layers: 1, 2, 4, 6
        \item Units: 32, 64, 128, 256, 512
    \end{enumerate}
    \item Transformer\begin{enumerate}
        \item Layers: 1, 2, 4, 6
        \item Embedding: 128, 256, 512, 1024
        \item Internal Dense Neurons: 256, 512, 1024
        \item Attention: 1, 2, 4, 8, 16
    \end{enumerate}
\end{enumerate}
\begin{figure}[tb]
    \centering
    \includegraphics[width=0.5\textwidth]{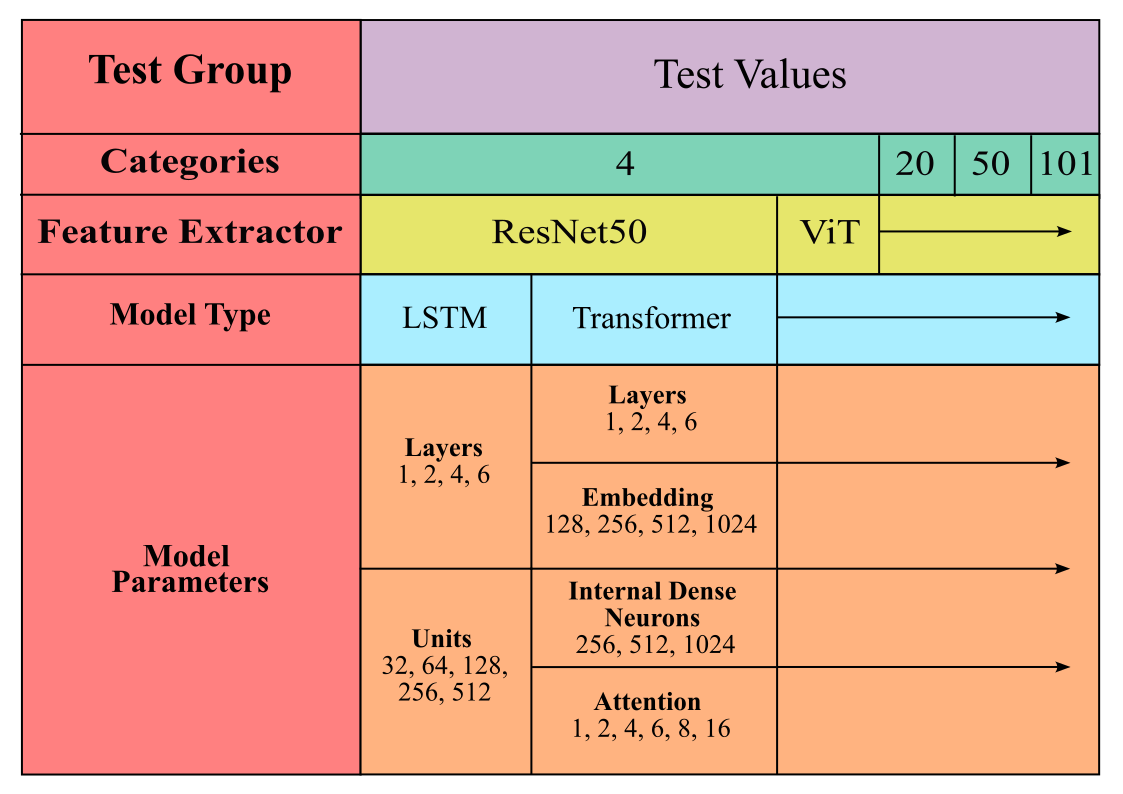}
    \caption{An overview of all tests performed.} \vspace{-5 mm}
    \label{fig:tests}
\end{figure}
\indent These tests are performed 8 times each. The tests are performed once for each type of feature extractor, and each feature extractor is used once for each different subset of the data. This means the LSTM will be tested 8 times (4 categories ResNet, 4 categories ViT, 20 categories ResNet, 20 categories ViT, 50 categories ResNet, 50 Categories ViT, Total dataset ResNet, and Total Dataset ViT), and the ReT will be tested 8 times in the same manner. This is done to find the model configurations with the best loss values.\\
\indent After, the best model configurations will be compared based upon their loss and accuracy on each of the data subsets to show accuracy on different dataset complexities. Then the training time and throughput time will be compared for the best configurations of the LSTM and ReT. Next, the throughput time of the ViT-Ret Model and the ResNet50-LSTM model will be compared. Finally, the memory usage of the two models will be compared.\\
\begin{figure*}[tb]
    \centering
    \includegraphics[width=0.82\textwidth]{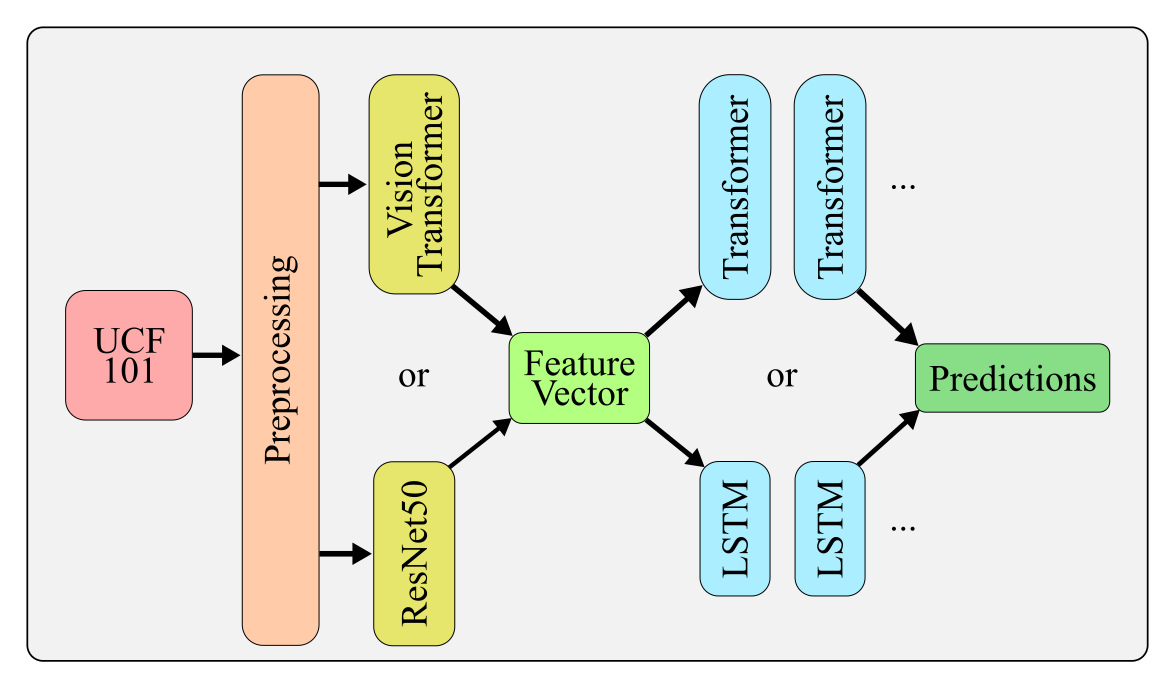}
    \caption{Model architectures.} 
    \label{fig:architectures}
\end{figure*}
\indent In our experiments, the model is trained using a shuffled 80 percent of the dataset. After each epoch of training, the training dataset is shuffled to help fight overtraining on the training dataset. For throughput testing, each model is tested by making predictions on the entire data subset at once. There are two important things to note about the testing process. First is the way the data is processed. Instead of creating a model that would perform the entire process end-to-end, each stage of model processing was handled separately. The data is preprocessed first and saved for later use. Secondly, neither the ReT nor ViT is implemented in parallel. This means that these tests do not utilize all the potential benefits of these models, and the models do not train or predict at the speeds that are possible if a parallel implementation is created. Parallel implementations of the transformer are the most obvious possible extension of this research which serves as a basis to show the viability of the transformer in activity recognition tasks.  The overall model flow for these tests can be seen in Fig. 17.

\section{Results} \label{sec:results}
We have split the experimental results into 5 sections. First, the accuracy of our models with the best loss values are show for each of the data subsets. Then, timing results are shown for the classification layers. Third, throughput times are shown comparing the ResNet50-LSTM model to the ViT-ReT model. Next, memory usage is discussed. Finally, the results from all sections are discussed together. We have performed all testing using the UCF 101 dataset. We have first converted the videos to sequences of 20 frames, with each frame resized to 224$\times$224 pixels. All training is performed with a batch size of 4 over 50 epochs.
\subsection{Accuracy} \label{subsec:accuracy}
For this section, we trained multiple different model configurations on different subsets of the UCF 101 dataset, and the configurations with the best loss values found are used for discussion. First, let us look at the models with the best loss values for the 4 Action dataset. The categorical cross entropy loss and accuracy for the optimal LSTM and Transformer model for each feature extractor is shown in Table \ref{tab:4actionaccuracy}. Best configurations:
\begin{itemize}
    \item ResNet-LSTM: 2 LSTM Layers, 256 Units
    \item ResNet-ReT: 1 ReT Layer, 128-Dim Embedding, 256 Neurons, 16 Attention Heads
    \item ViT-LSTM: 1 LSTM Layer, 256 Units
    \item ViT-ReT: 1 ReT Layer, 128-Dim Embedding, 256 Neurons, 8 Attention Heads
\end{itemize}
\begin{table}[t]
    \centering
    \caption{Accuracy with the 4 Action Database}
    \begin{tabular}{c|c|c|c|c}
        \hline
        \multirow{2}{5em}{4 Actions} & \multicolumn{2}{|c|}{ResNet50} & \multicolumn{2}{|c}{ViT}\\
        \cline{2-5}
        & \textit{Loss} & \textit{Accuracy} & \textit{Loss} & \textit{Accuracy} \\
        \hline
        LSTM & $4e^{-5}$ & 100\% & 0.50 & 81.25\% \\
        \hline
        ReT & $2e^{-4}$ & 100\% & 0.56 & 81.25\% \\
        \hline
    \end{tabular}
    \label{tab:4actionaccuracy}
\end{table}
\indent From these results we can see the LSTM and ReT models perform similarly on small datasets regardless of the feature extractor used. While the models using LSTM layers had slightly smaller loss values, there was no change in accuracy between the LSTM and ReT. There was, some loss in accuracy when switching from ResNet50 to the ViT, but the loss in accuracy is the same for both LSTM and ReT. Next, the results for the 20 Action dataset are shown in Table \ref{tab:20actionaccuracy}. Best configurations:
\begin{itemize}
    \item ResNet-LSTM: 1 Layer, 64 Units
    \item ResNet-ReT: 1 Layer, 128-Dim Embedding, 1024 Neurons, 16 Attention Heads
    \item ViT-LSTM: 2 Layers, 256 Units
    \item ViT-ReT: 1 Layer, 128-Dim Embedding, 256 Neurons, 8 Attention Heads
\end{itemize}
\indent Here, we begin to see the impact of increased dataset complexity and size. The loss of both models has increased by more than 100x their previous loss values when using ResNet50. However, this has not come with a significant decline in accuracy. The two models are still able to achieve close to 100\% accuracy with ResNet50. Interestingly, the loss of each model does not increase significantly, and the accuracy of each model does not decrease significantly from the smaller dataset when using the ViT, and even increases for the LSTM. Next, the results for the 50 Action dataset are shown in Table \ref{tab:50actionaccuracy}. Optimal configurations:
\begin{table}[t]
    \centering
    \caption{Accuracy with the 20 Action Database}
    \begin{tabular}{c|c|c|c|c}
        \hline
        \multirow{2}{5em}{20 Actions} & \multicolumn{2}{|c|}{ResNet50} & \multicolumn{2}{|c}{ViT}\\
        \cline{2-5}
        & \textit{Loss} & \textit{Accuracy} & \textit{Loss} & \textit{Accuracy} \\
        \hline
        LSTM & 0.06 & 98\% & 0.65 & 84.2\% \\
        \hline
        ReT & 0.06 & 100\% & 0.72 & 80.0\% \\
        \hline
    \end{tabular}
    \label{tab:20actionaccuracy}
\end{table}
\begin{table}[t]
    \centering
    \caption{Accuracy with the 50 Action Database}
    \begin{tabular}{c|c|c|c|c}
        \hline
        \multirow{2}{5em}{50 Actions} & \multicolumn{2}{|c|}{ResNet50} & \multicolumn{2}{|c}{ViT}\\
        \cline{2-5}
        & \textit{Loss} & \textit{Accuracy} & \textit{Loss} & \textit{Accuracy} \\
        \hline
        LSTM & 0.18 & 94.1\% & 1.31 & 73.6\% \\
        \hline
        ReT & 0.27 & 92.5\% & 1.33 & 73.8\% \\
        \hline
    \end{tabular}
    \label{tab:50actionaccuracy}
\end{table}
\begin{table}[t]
    \centering
    \caption{Accuracy with the 101 Action Database}
    \begin{tabular}{c|c|c|c|c}
        \hline
        \multirow{2}{5em}{101 Actions} & \multicolumn{2}{|c|}{ResNet50} & \multicolumn{2}{|c}{ViT}\\
        \cline{2-5}
        & \textit{Loss} & \textit{Accuracy} & \textit{Loss} & \textit{Accuracy} \\
        \hline
        LSTM & 0.33 & 93.2\% & 1.56 & 70.7\% \\
        \hline
        ReT & 0.39 & 93.2\% & 1.44 & 71.7\% \\
        \hline
    \end{tabular}
    \label{tab:101actionaccuracy}
\end{table}
\begin{itemize}
    \item ResNet-LSTM: 1 Layer, 128 Units
    \item ResNet-ReT: 1 Layer, 128-Dim Embedding, 1024 Neurons, 4 Attention Heads
    \item ViT-LSTM: 2 Layers, 512 Units
    \item ViT-ReT: 1 Layer, 128-Dim Embedding, 512 Neurons, 2 Attention Heads
\end{itemize}
\indent As should be expected, further increasing dataset complexity leads to a larger loss and lower accuracies for both models when given the same training configuration. However, the ReT and LSTM models are still performing very good with ResNet50 as a feature extractor. Even with 50 categories and over 6,500 total samples, the ResNet50-LSTM and ResNet50-ReT models are still able to maintain good accuracy. The accuracy of the ViT models is still lagging, however. Finally, the total dataset results are shown in Table \ref{tab:101actionaccuracy}. Optimal configuration:
\begin{itemize}
    \item ResNet-LSTM: 2 Layer, 64 Units
    \item ResNet-ReT: 1 Layer, 128-Dim Embedding, 1024 Neurons, 4 Attention Heads
    \item ViT-LSTM: 2 Layers, 512 Units
    \item ViT-ReT: 1 Layer, 128-Dim Embedding, 512 Neurons, 2 Attention Heads
\end{itemize}
\indent With the final increase in complexity, we continue to see the same trends as with the other datasets. The LSTM and ReT perform almost identically on all tasks. Their accuracy and loss values are so close that any difference could be due to different random starting weights. Their close performance maintains regardless of which features extractor is used. This serves to sufficiently prove that the Transformer can be used in place of the typical RNN while maintaining accuracy with optimal LSTM configurations and outperforming many others.\\
\indent As for the Vision Transformer vs ResNet50, the difference in performance is most likely due to its own pretraining. The vision transformer is most efficiently utilized when trained on a very large dataset and then finetuned. Most likely some combination of over-pooling and not using a sufficiently trained Vision Transformer resulted in the difference in accuracy here. While the vision transformer did not perform as well as ResNet50 in this case, its accuracy is still adequate for many tasks and shows promise with a better implementation.\\
\begin{table}[t]
    \centering
    \caption{Timing results with the 4 Actions Database}
    \begin{tabular}{c|c|c}
        \hline
        \multirow{2}{5em}{4 Actions} & \multicolumn{2}{|c}{Timing Results}\\
        \cline{2-3}
        & \textit{Training} & \textit{Throughput} \\
        \hline
        LSTM & 48.85s & 0.98s \\
        \hline
        ReT & 37.70s & 0.19s \\
        \hline
    \end{tabular}
    \label{tab:4actiontiming}
\end{table}
\begin{table}[t]
    \centering
    \caption{Timing results with the 20 Actions Database}
    \begin{tabular}{c|c|c}
        \hline
        \multirow{2}{5em}{20 Actions} & \multicolumn{2}{|c}{Timing Results}\\
        \cline{2-3}
        & \textit{Training} & \textit{Throughput} \\
        \hline
        LSTM & 2:37.00 & 0.51s \\
        \hline
        ReT & 3:04.68 & 0.80s \\
        \hline
    \end{tabular}
    \label{tab:20actiontiming}
\end{table}
\begin{table}[th!]
    \centering
    \caption{Timing results with the 50 Actions Database}
    \begin{tabular}{c|c|c}
        \hline
        \multirow{2}{5em}{50 Actions} & \multicolumn{2}{|c}{Timing Results}\\
        \cline{2-3}
        & \textit{Training} & \textit{Throughput} \\
        \hline
        LSTM & 6:57.90 & 0.80s \\
        \hline
        ReT & 8:16.60 & 0.80s \\
        \hline
    \end{tabular}
    \label{tab:50actiontiming}
\end{table}
\begin{table}[th!]
    \centering
    \caption{Timing results with the 101 Action Database}
    \begin{tabular}{c|c|c}
        \hline
        \multirow{2}{5em}{101 Action} & \multicolumn{2}{|c}{Timing Results}\\
        \cline{2-3}
        & \textit{Training} & \textit{Throughput} \\
        \hline
        LSTM & 21:01.75 & 1.72s \\
        \hline
        ReT & 16:16.01 & 1.42s \\
        \hline
    \end{tabular}
    \label{tab:101actiontiming}
\end{table}
\indent Now that it has been show that transformers perform with comparable accuracy to current methods, the next section will examine their training and prediction speeds.
\subsection{Classification Time} \label{subsec:classificationtime}
In this section, we show, for each UCF 101 subset, the timing data for the LSTM and ReT models, comparing their training time and throughput time for the entire data subset using our best model configuration. Of note, only timing data using ResNet50 will be shown. Since the model was implemented in a way that allows for direct timing of the LSTM or Transformer layers, it is not necessary to look at both feature extractors as the processing time of the Transformer and LSMT will be the same, or very close, for the different feature extractors. First, we will look at the timing data for the 4 Action subset. Timing data for this subset can be seen in Table \ref{tab:4actiontiming}.
\indent With this simple dataset, containing only 478 total videos and 4 categories, the ReT is faster than the LSTM. The ReT is able to train 1.30x faster than the LSTM models, and it is able to process the entire dataset 4.79x faster. This shows that, for smaller data subsets, the ReT model is significantly faster than the LSTM model. For the next data subset, timing data can be seen in Table \ref{tab:20actiontiming}.\\
\indent In the case of the slightly more complex data, 2500 videos with 20 action categories, the LSTM is marginally faster. The training of the ReT was 1.17x slower than the LSTM, and it processed the entire dataset 1.5x slower. Timing results for the 50-category dataset can be seen in Table \ref{tab:50actiontiming}:\\
\indent Again, as the complexity increases, 6,567 videos with 50 categories, the training time for the ReT begins to lag behind the training time of the LSTM, training 1.15x slower. However, once trained it is able to process the entire dataset in essentially the same amount of time. While it may take more time to train the ReT models to the same level as the LSTM models, once it is trained it runs just as fast as the LSTM models for a medium complexity dataset. Finally, the total dataset timing data is shown in Table \ref{tab:101actiontiming}:\\
\indent With the final and most complex dataset, 13,320 videos and 101 categories, the ReT performs much better than the LSTM model. It was able to train 1.29x faster, and able to process the entire dataset 1.21x faster. For the most complex dataset, the ReT is simply faster than using a RNN model. These results show the promise of the ReT for activity recognition tasks. The ReT performs just as accurately while being just as fast or faster than the LSTM model in most tasks. In the next section, a throughput analysis of the entire model structures is performed, comparing the ViT-ReT model to the traditional ResNet50-LSTM model.
\begin{table}[t]
    \centering
    \caption{Throughput timing results for different number of files}
    \begin{tabular}{c|c|c|c|c}
        \hline
        \multirow{2}{*}{Throughput} & \multicolumn{4}{|c}{Files Predicted}\\
        \cline{2-5}
        & \textit{1} & \textit{100} & \textit{1000} & \textit{10000}\\
        \hline
        \makecell{ViT-\\ReT} & 1.32s &  4.07s & 35.77s & 5:59.25\\
        \hline
        \makecell{ResNet50-\\LSTM} & 2.69s & 9.38s & 1:10.43 & 11:52.03 \\
        \hline
    \end{tabular}
    \label{tab:throughput}
\end{table}
\begin{table}[tb]
    \centering
    \caption{Throughput fps results for different number of files}
    \begin{tabular}{c|c|c|c|c}
        \hline
        \multirow{2}{*}{FPS} & \multicolumn{4}{|c}{Files Predicted}\\
        \cline{2-5}
        & \textit{1} & \textit{100} & \textit{1000} & \textit{10000}\\
        \hline
        \makecell{ViT-\\ReT} & 15.15 &  491 & 559 & 557\\
        \hline
        \makecell{ResNet50-\\LSTM} & 7.43 & 213 & 283 & 281 \\
        \hline
        \makecell{ViT-ReT\\Speedup} & 2x & 2.24x & 1.98x & 1.98x \\
        \hline
    \end{tabular}
    \label{tab:fps}
\end{table}
\begin{table}[th!]
    \centering
    \caption{Memory usage of ViT-ReT vs ResNet50-LSTM}
    \begin{tabular}{c|c|c|c|c}
        \hline
        \multirow{2}{*}{\makecell{Memory\\(MB)}} & \multicolumn{4}{|c}{Action Categories}\\
        \cline{2-5}
        & \textit{4} & \textit{20} & \textit{50} & \textit{101}\\
        \hline
        \makecell{ViT-\\ReT} & 1021.15 & 1019.28 & 1015.84 & 1021.66\\
        \hline
        \makecell{ResNet50\\-LSTM} & 1121.42 & 1122.46 & 1117.06 & 1121.5\\
        \hline
        \makecell{Memory\\Efficiency\\Gained} & 1.10x & 1.10x & 1.10x & 1.10x\\
        \hline
    \end{tabular}
    \label{tab:mem}
\end{table}
\subsection{Total Throughput Time} \label{subsec:totalthroughputtime}
To measure model throughput, our best performing ReT and LSTM configurations for the total dataset are used. Once trained, the ViT is used for feature extraction with the ReT and ResNet50 is used for the feature extraction with the LSTM (ViT-ReT and ResNet50-LSTM). These models are then passed a randomly selected subset of the training data to make predictions. The time for each prediction was tracked and used to calculate throughput. Each video file contains 20 frames. Time data is shown in Table \ref{tab:throughput}.\\ 
\indent As shown above, the ViT-ReT model is significantly faster than ResNet50-LSTM model. In most case, the ViT-ReT is 2x faster than the ResNet50-LSTM model. This speedup can be seen in Table \ref{tab:fps}, along with the fps for the two models with each different number of files.\\
\indent These results clearly show that the ViT-ReT model is significantly faster, in all cases approximately 2x faster, than the ResNet50-LSTM model. This is a clear example of the promise transformers have in activity recognition. Regardless of the complexity of the dataset, the ViT-ReT is faster. It is important to note that, in the case of the entire model flow, the ResNet50-LSTM model is still more accurate than the ViT-ReT model, but with a better implementation of the ViT, this accuracy disparity would most likely be overcome. In the next section, the memory used in the creation of both models is compared.
\subsection{Memory} \label{subsec:memory}
Since it has been shown that the ReT is just as accurate while being faster than the LSTM, the next metric to consider is the memory usage of the two different models. To track this, the process’s memory usage was tracked before and after creation of the models using the best model configuration for each data subset. The difference in the memory usage measured before and after each task shows how much memory performing a certain action with a specific model will take. The memory data can be seen in Table \ref{tab:mem}.\\
\indent As is shown, the ViT-ReT model is more memory efficient than the ResNet50-LSTM model while still being faster. On average, the transformer model is 1.1x more memory efficient than the ResNet50-LSTM model. Even using multiple different model configurations, the transformer models are more memory efficient. In the next section, our results from all testing sections will be discussed in detail.  
\subsection{Discussion} \label{subsec:discussion}
The most striking result is the ability of the ReT model to maintain similar accuracy with the LSTM model while being faster and more memory efficient. In all cases, the ReT model exhibits comparable cross entropy loss and accuracy to the LSTM model. This makes TNN models ideal not only for general use, but also for use in edge devices with limited resources. The transformer, even if not implemented in parallel, can use resources more efficiently by using less memory and by using that memory for less time than a traditional RNN activity recognition model without any significant drop in performance. Clearly, an implementation that utilizes any amount of parallelization to perform more quick matrix multiplication (the backbone of multi-headed attention) would allow for even better use of resources.  Overall, the transformer has shown itself to be a better and more versatile model than the traditional RNNs when used for activity recognition.\\
\indent For the ViT model, it maintains decent accuracy, but was significantly behind its deep CNN counterpart in accuracy. This is most likely due to the specific implementation of the ViT used and the amount of training performed for it. As shown by Dosovitskiy et al. \cite{Dosovitskiy16Words}, it can perform comparable to other state-of-the-art deep CNN models, and even more efficiently in some cases. With a better implementation it is still possible to achieve accuracies close to or better than state-of-the-art deep CNNs for activity recognition. However, the ViT, when used in conjunction with the ReT, creates a model that is faster and more memory efficient than a traditional model using ResNet50 and LSTM classifiers. This serves to show that, with a better trained implementation, there is a potential for the ViT to make significant improvements over current deep CNNs.

\section{Conclusion} \label{sec:conclusions}
Activity recognition is a challenging research area in computer vision that involves recognizing the actions taken by humans with different sensors. Contemporary activity recognition models rely on multiple very dense and complex networks which make these models unsuitable for resource-constrained edge devices. Deep residual CNNs and RNN chains, the backbone of modern activity recognition models, suffer from complexity and speed issues, making real-time applications or computations in limited resource environments difficult. There is a need to advance activity recognition models beyond what currently exists, and the transformer neural networks (TNNs) provide a propitious alternative.\\
\indent The TNNs are one of the most promising neural networks created in the recent years. TNNs have revolutionized sequence-to-sequence modeling and shown that it is possible to create very accurate and lightweight models with little more than matrix multiplications and fully connected artificial neural networks. This paper sought to show the potential applications of TNNs beyond sequence-to-sequence models. It was also shown that there is potential for Transformers to create significant improvements in computer vision and activity recognition by replacing both sets of complex networks that models currently rely on. The recurrent transformer (ReT), and its extension the vision transformer (ViT), show promise in these domains, with the potential to significantly improve on current state-of-the-art models. Their application into nonsequential and non-time dependent datasets could prove to make huge improvements in a wide variety of machine learning tasks.\\
\indent Our results have shown that the ReT can make predictions that maintain similar accuracy to traditional RNN classifiers used for activity recognition while being faster and more memory efficient. When used in conjunction with a ViT for feature extraction, there is a significant speedup over current deep CNN and RNN activity recognition models, as well as a notable improvement in memory efficiency. This proves the potential utility of TNNs for activity recognition.
\section{Future Research Directions} \label{sec:futureresearch}
There are three immediate directions for future research that extends this paper. The first is creating a more accurate Vision Transformer and completing Vision Transformer Focused testing to show its application to the activity recognition Field. This would show without a doubt the utility of the Transformer in all fields, not only sequence-to-sequence problem fields. It can also serve to significantly improve the model efficiency of activity recognition models. With a better vision transformer implementation, testing should focus on speed and memory efficiency while maintaining accuracy, as is done in this paper. The tests discussed in this paper can then be extended to including training and testing utilizing the entire activity recognition model end to end including all preprocessing and feature extraction to show the overall improvements made to the activity recognition task by both the recurrent and vision transformer models.\\
\indent Second, creating fully parallel implementation of both the recurrent and vision transformer networks would show how powerful TNNs can be when they are able to fully utilize the maximum potential of their architecture. TNNs are already faster and more memory efficient than the RNN networks currently in use without a parallel implementation that makes use of fast matrix multiplication to drastically increase model speed. With better parallel implementations, TNN activity recognition models can be used in many different resource-constrained environments and/or real-time systems.\\
\indent Finally, we believe it may be possible to create a TNN for activity recognition that performs both feature extraction and classification within a single model, removing the need for a deep CNN or ViT altogether. This would be accomplished by using the embedding layers to perform a pseudo-feature extraction before data processing begins in the ReT. With the right embedding and preprocessing steps, it may be possible to achieve similar accuracy without using CNNs, instead using what may be a significantly more lightweight and efficient overall model implementation. This could potentially lead to creating the fastest and most memory-efficient activity recognition model in existence.

\section*{Acknowledgments}
The authors would like to acknowledge Erik Blasch for his involvement in the concept development of this work. This research was supported in part by the Air Force Office of Scientific Research (AFOSR) Contract Number FA9550-22-1-0040. Any opinions, findings, and conclusions or recommendations expressed in this material are those of the author(s) and do not necessarily reflect the views of the Air Force, AFRL, and/or AFOSR.

\section*{Appendix}
Code used to create the models that served as the basis for this paper can be found at: \url{https://github.com/JamesWensel/TranformerActivityRecognition}

\bibliographystyle{IEEEtran}
\bibliography{References.bib}


\end{document}